\documentclass{article}

\usepackage[T1]{fontenc}
\usepackage{microtype}
\usepackage{graphicx}
\usepackage{subcaption}
\usepackage{booktabs}
\usepackage{hyperref}

\usepackage[preprint]{icml2026}

\usepackage{amsmath}
\usepackage{amssymb}
\usepackage{xcolor}
\usepackage{listings}
\usepackage{enumitem}
\setlist[enumerate,itemize]{topsep=0pt,itemsep=0pt,leftmargin=18pt}

\newcommand{\papertitle}{Transcoder Adapters for Reasoning-Model Diffing}

\icmltitlerunning{\papertitle}

\begin{document}

\twocolumn[
  \icmltitle{\papertitle}

  \icmlsetsymbol{equal}{*}  

  \begin{icmlauthorlist}
    \icmlauthor{Nathan Hu}{aff1}
    \icmlauthor{Jake Ward}{aff2}   
    \icmlauthor{Thomas Icard}{aff1}  
    \icmlauthor{Christopher Potts}{aff1}
  \end{icmlauthorlist}

  \icmlaffiliation{aff1}{Stanford University}
  \icmlaffiliation{aff2}{MATS}

  \icmlcorrespondingauthor{Nathan Hu}{nathu@cs.stanford.edu}

  \icmlkeywords{Machine Learning, ICML}

  \vskip 0.3in
]

\printAffiliationsAndNotice{}

\begin{abstract}
While reasoning models are increasingly ubiquitous, the effects of reasoning training on a model's internal mechanisms remain poorly understood.  
In this work, we introduce transcoder adapters, a technique for learning an interpretable approximation of the difference in MLP computation before and after fine-tuning.
We apply transcoder adapters to characterize the differences between Qwen2.5-Math-7B and its reasoning-distilled variant, DeepSeek-R1-Distill-Qwen-7B. 
Learned adapters are faithful to the target model's internal computation and next-token predictions. When evaluated on reasoning benchmarks, adapters match the reasoning model's response lengths and typically recover 50–90\% of the accuracy gains from reasoning fine-tuning. Adapter features are sparsely activating and interpretable. When examining adapter features, we find that only ~8\% have activating examples directly related to reasoning behaviors. 
We deeply study one such behavior---the production of hesitation tokens (e.g., `wait'). Using attribution graphs, we trace hesitation to only ~2.4\% of adapter features (5.6k total) performing one of two functions. These features are necessary and sufficient for producing hesitation tokens; removing them reduces response length, often without affecting accuracy. Overall, our results provide insight into reasoning training and suggest transcoder adapters may be useful for studying fine-tuning more broadly.%
\ificmlshowauthors{} See \url{https://transcoder-adapters.github.io/} for code, checkpoints, and interactive visualizations.\fi
\end{abstract}

\section{Introduction}
\label{sec:introduction}

\begin{figure}[t]
    \centering
    \includegraphics[width=\linewidth]{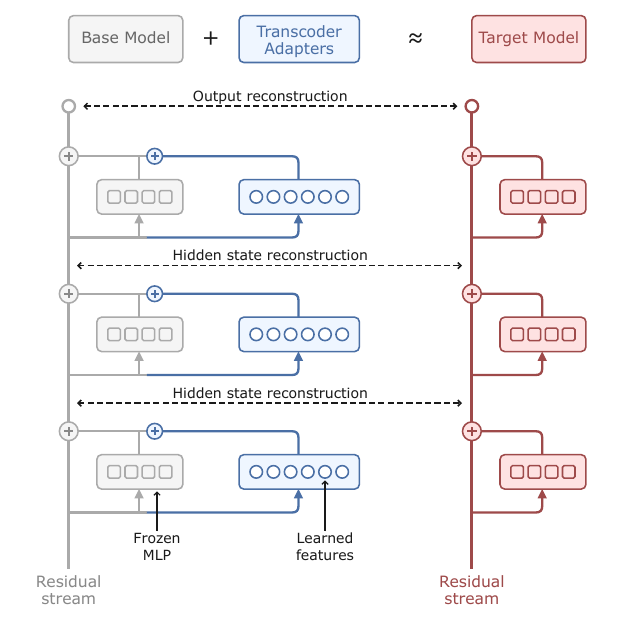}
    \caption{\textbf{Transcoder adapters for model diffing.} Transcoder adapters learn a sparse approximation of the difference in MLP computation before and after fine-tuning. Adapters are trained to reconstruct both internal activations and final output. We apply transcoder adapters to characterize the differences between Qwen2.5-Math-7B and its reasoning-distilled variant, DeepSeek-R1-Distill-Qwen-7B.}
    \label{fig:method}
\end{figure}

Reasoning models are increasingly ubiquitous: nearly all current frontier language models, including GPT-5 \citep{singh2025openaigpt5card}, Claude Opus 4.5 \citep{anthropic2025claude}, DeepSeek-R1 \citep{Guo_2025}, and Gemini 2.5 \citep{comanici2025gemini25pushingfrontier}, are capable of an extended thinking mode. Such reasoning responses are characterized by long sequences of intermediate reasoning tokens and higher final-answer accuracy.
Despite this widespread adoption, fully understanding the effects of reasoning training remains an active area of research \citep{yue2025doesreinforcementlearningreally, gandhi2025cognitivebehaviorsenableselfimproving,wu2026invisibleleashrlvrescape}. We approach this question from an interpretability perspective and examine the effects of reasoning training on a model's internal mechanisms.


We introduce transcoder adapters, which learn a sparse approximation of the difference in MLP computation between a base model and its fine-tuned variant (Figure~\ref{fig:method}). At each layer, a transcoder adapter runs in parallel with the frozen base model MLP. Adapters are trained to reconstruct the target model's internal activations and final output. Directly modeling the difference in computation deviates from typical sparse dictionary learning methods, which aim to fully reconstruct model representations or computation. This design decision trades insight into the base model's computation for more targeted analysis of fine-tuning. We find that the difference in MLP computation is a far easier object to decompose and study than the full MLP. Adapters achieve faithful reconstruction with an order of magnitude fewer active features than typical transcoders\footnote{Our adapters achieve L0 of 0.1–10, compared to 20–200 typical for full reconstruction \citep{karvonen2025saebenchcomprehensivebenchmarksparse}.}, and the resulting model remains coherent when generating thousands of tokens. Additionally, each adapter feature directly reflects a change induced by fine-tuning.\footnote{In contrast with methods like crosscoders \citep{lindsey2024crosscoders}, which jointly learn features for both models, then identify model specific features.}

We apply transcoder adapters to characterize the differences between Qwen2.5-Math-7B \citep{yang2024qwen25mathtechnicalreportmathematical} and its reasoning-distilled variant, DeepSeek-R1-Distill-Qwen-7B \citep{Guo_2025}, and verify the quality of trained adapters. Adapters are faithful to the target model's internal computation and next-token predictions. 
When evaluated on reasoning benchmarks, adapters match the reasoning model’s response lengths and typically recover 50–90\% of the accuracy gains from reasoning
fine-tuning. Adapter features are sparsely activating and interpretable, achieving higher automated interpretability scores than MLP neurons.
  
We lastly study the effects of study reasoning fine-tuning by interpreting transcoder adapters. When classifying features based on activating text examples, we find that only ~8\% of features appear directly related to reasoning behaviors. Far more appear related to domain knowledge or general language modeling. This rough measure suggests more of reasoning training's effect may stem from increasing domain knowledge rather than learning reasoning-specific structure. We deeply study one reasoning behavior---the production of hesitation tokens (e.g., `wait'). Using attribution graphs, we trace hesitation to 2.4\% of adapter features (5.6k total) performing one of two functions. These features are necessary and sufficient for hesitation; removing them from the adapter reduces response length by over 50\% with no decrease in accuracy on three of four benchmarks. We confirm that suppressing these adapter features in the target reasoning model itself also reduces response length, with only a slight decrease in accuracy.


\textbf{Contributions.} Our main contributions are: (a) the introduction and validation of transcoder adapters, a technique for learning an interpretable approximation of the difference in MLP computation before and after fine-tuning, and (b) two mechanistic insights from our case study of reasoning fine-tuning: a broad characterization of learned differences and a detailed account of hesitation behavior.
\section{Related Work}
\label{sec:related_work}
\textbf{Sparse dictionary learning for LLM interpretability.} Transcoder adapters build on sparse dictionary learning methods.
The most prevalent of these approaches are sparse autoencoders (SAEs), which are trained to reconstruct model activations and have been shown to reveal interpretable directions in activation space \citep{cunningham2023highlyinterpretable, templeton2024scaling, gao2024scalingevaluating}. Our approach builds upon three extensions of SAEs: transcoders, which model MLP computation rather than reconstructing activations \citep{dunefsky2024transcoders, ameisen2025circuit}; crosscoders, which identify shared and unique features when comparing the representations of two models \citep{lindsey2024crosscoders, minder2025overcomingsparsityartifactscrosscoders, jiralerspong2025crossarchitecture}; and work directly training SAEs on activation differences \citep{aranguri2025saeactivationdiff, dumas2025diffbasechat}.

\textbf{Reasoning model interpretability.} Closely related to our work, \citet{baek2025understandingdistilledreasoningmodels} and \citet{troitskii2025internalstateswaitmodulate} train crosscoders to compare base and reasoning model representations at select layers, finding features that elicit certain reasoning behaviors when used for steering. Our work introduces transcoder adapters to pursue a similar line of inquiry. Beyond sparse dictionary learning approaches, there is a much broader field of reasoning model interpretability including methods such as resampling rollouts \citep{macar2025thoughtbranchesinterpretingllm}, probing model representations \citep{zhang2025reasoningmodelsknowtheyre}, and analysis of attention heads \citep{zhang2025reasoninganswerempiricalattentionbased}. 

\textbf{Ease of reasoning elicitation.} Recent work has sought to understand reasoning behavior by studying minimal interventions that elicit reasoning from non-reasoning models. These interventions include adapting few parameters via low-rank LoRAs \citep{ward2025rank1lorasencodeinterpretable, schulman2025lora}, modifying only specific parameters such as layerwise biases \citep{sinii2025steeringllmreasoningbiasonly} or specific modules \citep{shao2025reasonslargelanguagemodels}, selectively applying steering vectors \citep{venhoff2025basemodelsknowreason}, and fine-tuning on as few as 1000 reasoning examples \citep{muennighoff2025s1simpletesttimescaling}. Other work shows that characteristic self-reflection behaviors of reasoning models are already exhibited by base models \citep{liu2025understandingr1zeroliketrainingcritical}. 

\textbf{Mechanistic studies of fine-tuning.} Several lines of work study how fine-tuning affects models. At the mechanistic level, some find that fine-tuning reinforces existing mechanisms in the model \citep{prakash2024finetuningenhancesexistingmechanisms} or learns simple wrappers around prexisting capabilities \citep{jain2024mechanisticallyanalyzingeffectsfinetuning}. Mechanistic studies of specific tasks find that instruction tuning modifies attention and MLP weights to orient toward user tasks \citep{wu2024languagemodelinginstructionfollowing}, while safety fine-tuning minimally transforms MLP weights to cluster safe and unsafe activations separately \citep{jain2024makesbreakssafetyfinetuning}. 

\section{Training Transcoder Adapters}
\label{sec:training}


\subsection{Architecture}

Transcoder adapters follow the standard transcoder architecture \citep{dunefsky2024transcoders, ameisen2025circuit}. At each layer $\ell$ of the target model, we learn a transcoder $T^\ell$ with encoder parameters $W_\text{enc}^\ell \in \mathbb{R}^{d_\text{features}\times d_\text{model}}$, $b_\text{enc}^\ell \in \mathbb{R}^{d_\text{features}}$, and decoder parameters $W_\text{dec}^\ell \in \mathbb{R}^{d_\text{model} \times d_\text{features}}$, $b_\text{dec}^\ell \in \mathbb{R}^{d_\text{model}}$. Letting $x \in \mathbb{R}^{d_\text{model}}$ denote the input to the transcoder, the transcoder computes feature activations:
\[
    a^\ell(x) = \text{ReLU}(W_\text{enc}^\ell x + b_\text{enc}^\ell)
\]
and output:
\[
    T^\ell(x) = W_\text{dec}^\ell a^\ell(x) + b_\text{dec}^\ell
\]

Each of the $d_\text{features}$ rows of $W_\text{enc}^\ell$ and columns of $W_\text{dec}^\ell$ corresponds to a learned feature. The $i$-th feature is considered \emph{active} for input $x$ if its activation $a^{\ell}_i(x) > 0$.

Traditional transcoders are trained to fully replace an MLP's computation. We instead train adapters to approximate the difference between base and target model MLPs, such that $\text{MLP}_\text{base}^\ell(x) + T^\ell(x) \approx \text{MLP}_\text{target}^\ell(x)$. More precisely, we build a \emph{replacement model} by replacing each MLP in the target model with a base MLP and transcoder adapter. At each layer, the replacement model forward pass computes:
\begin{align*}
    \hat{h}_\ell' &= \hat{h}_{\ell-1} + \text{Attn}_\text{target}^\ell(\hat{h}_{\ell-1}) \\
    \hat{h}_\ell &= \hat{h}_\ell' + \text{MLP}_\text{base}^\ell(\hat{h}_\ell') + T^\ell(\hat{h}_\ell')
\end{align*}
where $\hat{h}_0$ is the embedded input and $h_\ell$ and $\hat{h}_\ell$ denote hidden states at layer $\ell$ for the target and replacement models. Finally, we let $y_{\text{target}}$ and $\hat{y}$ denote the output distributions of the target and replacement models respectively. Only transcoder parameters are updated during training.

\subsection{Training Objective}
\label{sec:training_losses}
We train transcoder adapters to faithfully reconstruct the target model's outputs and hidden states. In addition to faithful reconstruction, adapter features must be sparsely activating.

For \textbf{output reconstruction}, we penalize KL divergence between output distributions:
\[
    \mathcal{L}_\text{KL} = \text{KL}(y_\text{target}, \hat{y})
\]

For \textbf{hidden state reconstruction}, we use two loss terms. The first is the normalized MSE between activations:
\[
    \mathcal{L}_\text{NMSE} = \sum_\ell \frac{\| \hat{h}_\ell - h_\ell \|_2^2}{\| h_\ell \|_2^2}
\]
This is the standard L2 loss used in SAE and transcoder training \citep{cunningham2023highlyinterpretable, templeton2024scaling}.

We additionally incentivize hidden state reconstruction using a \textit{bridging loss}, which measures KL divergence when feeding activations from one model through the remaining layers of the other. Let $M^\ell$ denote the $\ell$-th transformer block of the target model and $\hat{M}^\ell$ the corresponding replacement block (with base MLP and learned transcoder). The bridging losses are:
\begin{align*}
    \mathcal{L}_\text{bridge}^{r \to t} &= \sum_{\ell=1}^{L} \text{KL}\left( y_\text{target},\, (M^L \circ \cdots \circ M^{\ell+1})(\hat{h}_\ell) \right) \\
    \mathcal{L}_\text{bridge}^{t \to r} &= \sum_{\ell=1}^{L} \text{KL}\left( y_\text{target},\, (\hat{M}^L \circ \cdots \circ \hat{M}^{\ell+1})(h_\ell) \right)
\end{align*}
These losses can be viewed as a natural extension of end-to-end SAE training \citep{braun2024identifyingfunctionallyimportantfeatures} to the multilayer setting. Because we learn a transcoder at each layer, there are $2^L$ ways to combine sparse layers and true layers for end-to-end reconstruction. The bridging loss is a first-order approximation of this, proposed by \citet{gao2025weightsparsetransformersinterpretablecircuits} for bridging between dense and weight-sparse transformers.

For \textbf{sparsity}, we penalize the L1 norm of feature activations weighted by decoder norm, as is typical in SAE and transcoder training \citep{dunefsky2024transcoders}:
\[
    \mathcal{L}_\text{sparsity} = \sum_{\ell=1}^{L} \sum_{i=1}^{d_\text{features}} \| W_{\text{dec},i}^\ell \|_2 \cdot a_i^\ell
\]
where $W_{\text{dec},i}^\ell$ is the $i$-th decoder column at layer $\ell$.

\subsection{Experimental Setup}

We train transcoder adapters to approximate the difference between Qwen2.5-Math-7B and DeepSeek-R1-Distill-Qwen-7B. For simplicity, we refer to these as the base model and target model respectively. 
We learn a transcoder with 8192 features at each of the model's 28 layers. We train on 50k samples ($\sim$380M tokens) from the OpenThoughts3 dataset \citep{guha2025openthoughtsdatarecipesreasoning}, a dataset of curated reasoning transcripts from QwQ-32B \citep{qwq32b}. Additional training and dataset details are provided in Appendix~\ref{sec:training_details}.

\section{Evaluating Transcoder Adapters}
\label{sec:evaluation}

We train 5 transcoder adapters with $L_0$ ranging from 0.1 to 10, where $L_0$ is the number of active features averaged across layers and tokens. We evaluate adapters on output faithfulness, internal faithfulness, and benchmark performance. We compare performance against two baselines: Qwen2.5-Math-7B (the \textbf{base model}) and a \textbf{hybrid model} constructed by replacing all MLP parameters in R1-Distill-Qwen-7B with base model parameters. A key limitation of transcoder adapters is that they only capture differences in MLP parameters. The hybrid baseline is essential to verify that differences in non-MLP parameters (attention, embeddings) do not already account for the reasoning behavior we wish to study. To confirm we are not under-eliciting the hybrid baseline, in Appendix~\ref{sec:hybrid_appendix}, we additionally explore refitting RMSNorm parameters and few-shot prompting during evaluations, but find no method that significantly improves performance.

R1-Distill-Qwen-7B was trained on 800k samples from DeepSeek-R1 \citep{Guo_2025}, while we train transcoder adapters on 50k samples from QwQ-32B. We expect some gap in faithfulness to be inevitable given these training data differences. To bound this error, we consider an \textbf{MLP fine-tuning skyline}. Starting from the hybrid model, we fine-tune only the MLPs to minimize KL divergence against the target model. 

\subsection{Output Faithfulness}
\begin{figure}[t]
    \centering
    \includegraphics[width=0.9\linewidth, trim=0.3cm 0 0 0, clip]{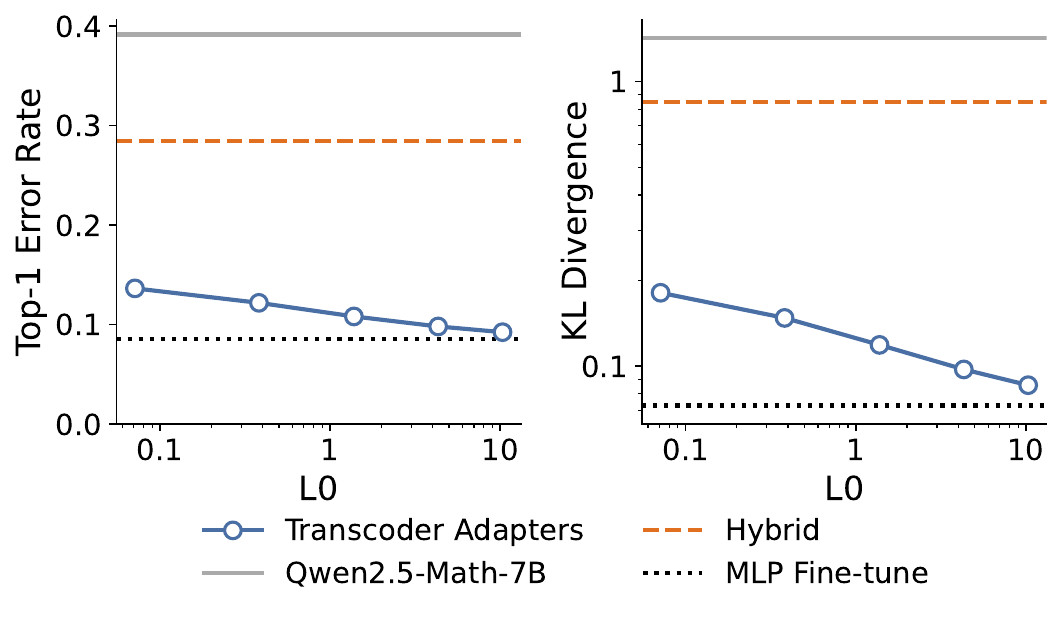}
    \caption{\textbf{Output faithfulness.} Top-1 error and KL divergence against the target model. Transcoder adapters outperform baselines and approach the MLP fine-tuning skyline, achieving strong reconstruction even at very low sparsity ($L_0$ of 0.1–10).}
    \label{fig:token_recon}
\end{figure}

We first evaluate how well transcoder adapters reconstruct the target model's next-token predictions. We measure top-1 error and KL divergence between output distributions at each token position over 5M tokens sampled from R1-Distill-Qwen-7B (Figure~\ref{fig:token_recon}). Because many tokens are low entropy, even the base model agrees with the target on most predictions (39.2\% top-1 error). The hybrid baseline reduces top-1 error to 28.5\%, but a substantial gap remains despite sharing all non-MLP parameters with the target. Transcoder adapters close this gap: even the sparsest, at $L_0$ of 0.1, achieves 13.6\% top-1 error, decreasing to 9.2\% at $L_0$ of 10. This approaches the MLP fine-tuning skyline (8.5\%). KL divergence follows the same pattern. 




\subsection{Internal Faithfulness}

\begin{figure}[t]
    \centering
    \includegraphics[width=\linewidth]{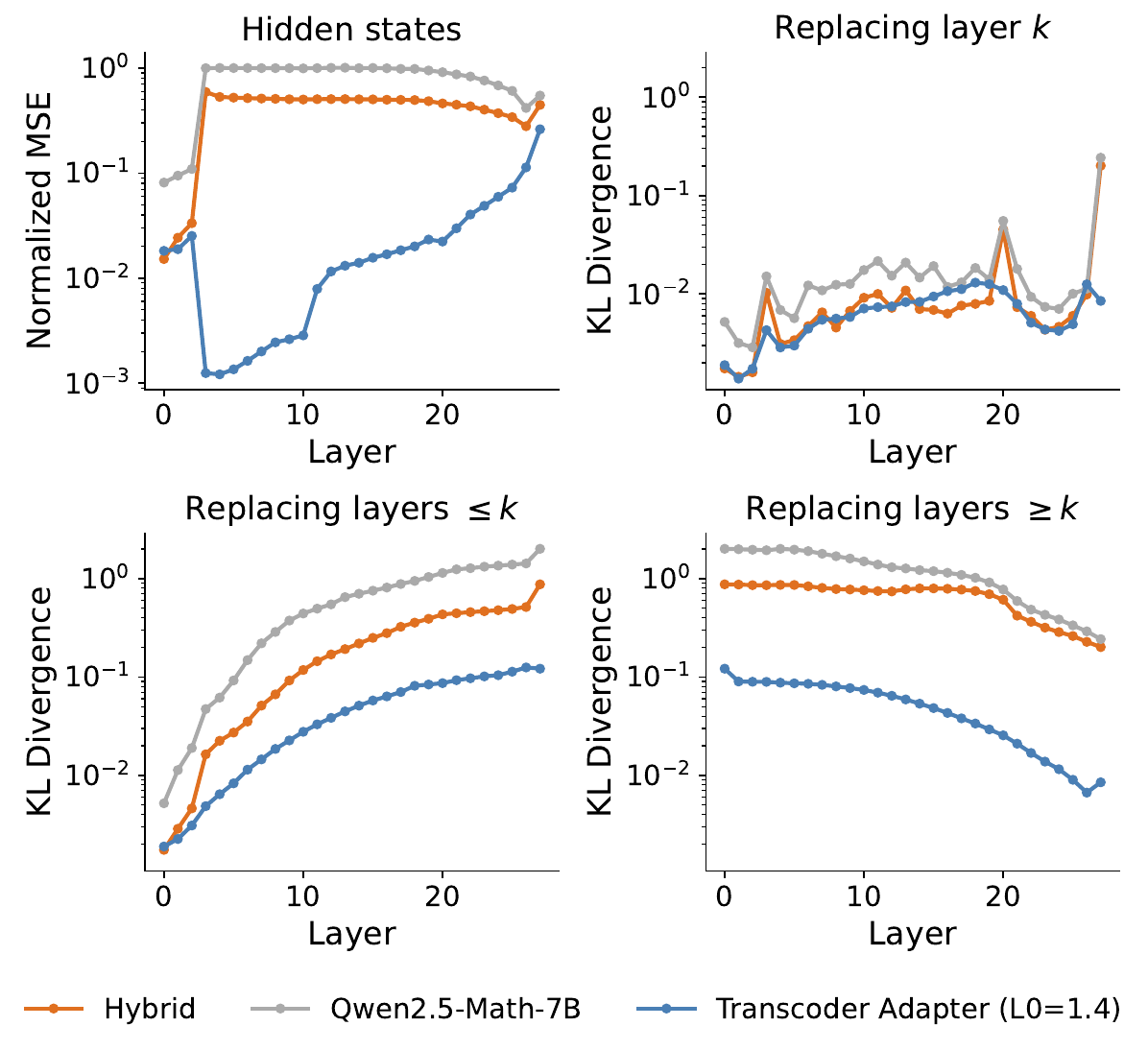}
    \caption{\textbf{Internal faithfulness.} We evaluate internal faithfulness via (1) NMSE of hidden states against the target model and (2) KL divergence when replacing various subsets of target layers with adapter layers. Transcoder adapters outperform baselines across all metrics. Notably, KL divergence when replacing subsets of layers is always less than KL divergence when using adapters to approximate all layers, indicating \textit{internal errors do not accumulate beyond what is reflected in the final output}.}
    \label{fig:internal_metrics}
\end{figure}


To evaluate internal faithfulness, we measure normalized MSE between adapter and target hidden states. While NMSE captures raw reconstruction error, it is hard to interpret in a vacuum. To contextualize this, we perform partial replacements—substituting the first $k$, final $k$, or a single layer $k$ with adapter layers—and measure output KL divergence. 
Notably, KL divergence when replacing subsets of layers is always less than KL divergence when using adapters to approximate all layers, indicating that internal errors do not accumulate beyond what is reflected in the final output.
Results for $L_0$ 1.4 adapter are shown in Figure~\ref{fig:internal_metrics}. Appendix~\ref{sec:faithfulness_appendix} shows these metrics for our full suite of adapters and for an adapter trained without the bridging loss. When ablating the bridging loss, NMSE remains low but KL divergence under partial replacement increases substantially. Transcoder adapters outperform baselines on all metrics. Divergence decreases monotonically as more target layers are used, suggesting adapter layers independently approximate their target layer's computation rather than relying on later layers to compensate for earlier errors. 

\subsection{Benchmark Evaluation}

\begin{figure}[t]
    \centering
    \includegraphics[width=0.95\linewidth]{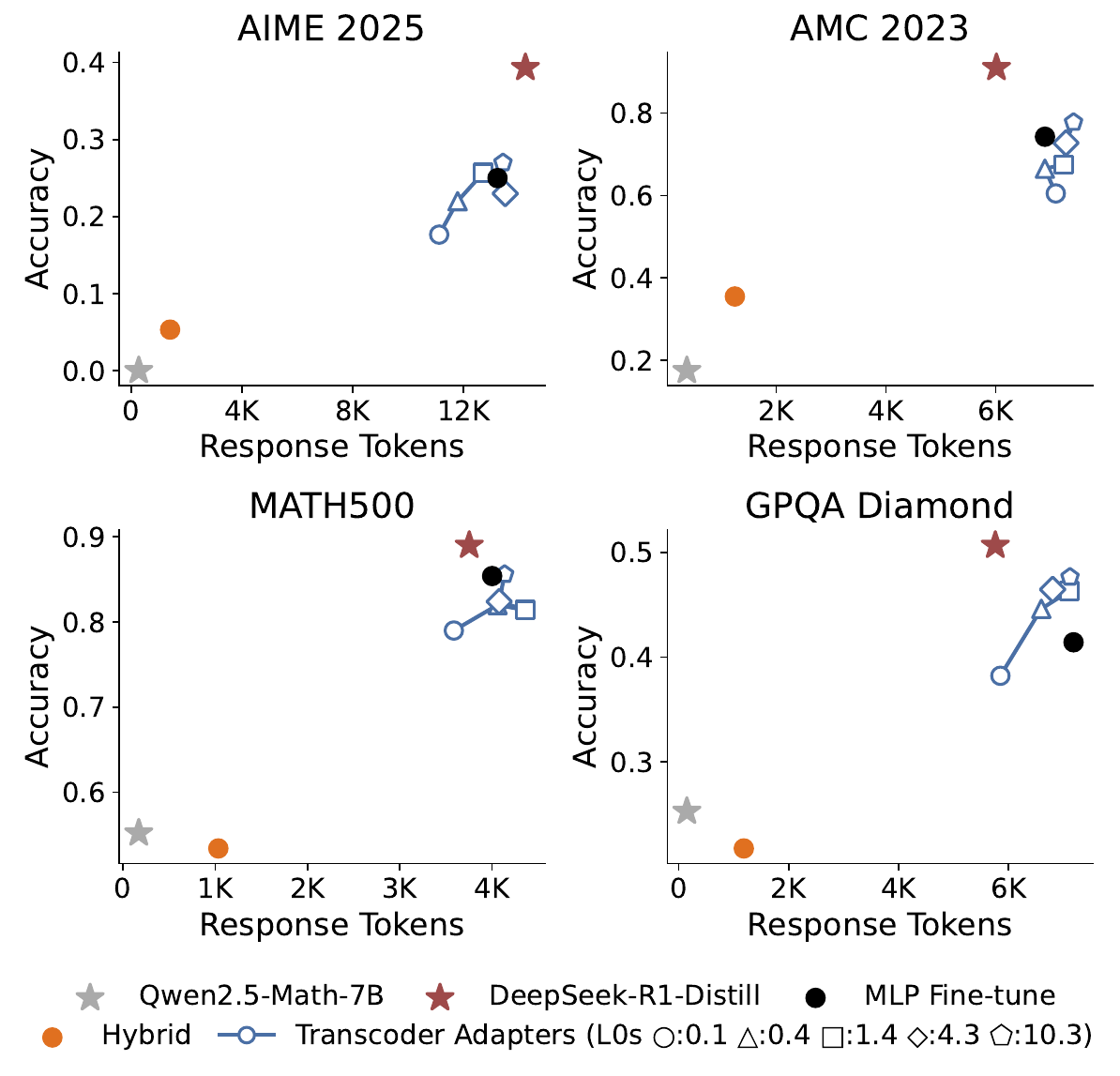}
    \caption{\textbf{Benchmark evaluation.} Transcoder adapters match the target model's response lengths and recover much of the accuracy gains from reasoning fine-tuning. The remaining gap is comparable to the MLP fine-tuning skyline, suggesting it reflects training data differences rather than limitations of transcoder adapters. Despite sharing all non-MLP parameters with the target reasoning model, the hybrid baseline exhibits similar response length and accuracy to the base model.}
    \label{fig:benchmark_eval}
\end{figure}

Taking advantage of the fact that transcoder adapters faithfully approximate the target model and produce coherent outputs, we evaluate replacement models on standard reasoning benchmarks: MATH500 \citep{hendrycks2021measuringmathematicalproblemsolving}, AMC23 \citep{amc2023}, AIME25 \citep{aime2025}, and GPQA Diamond \citep{rein2023gpqagraduatelevelgoogleproofqa}. We use Evalchemy \citep{evalchemy2025} as our evaluation engine; Appendix~\ref{sec:eval_details} contains additional evaluation details.

Figure~\ref{fig:benchmark_eval} summarizes these benchmark evaluations. Transcoder adapters recover a large fraction of the accuracy gains from reasoning fine-tuning. Adapters exhibit the characteristic long response traces of reasoning models, with response lengths varying across benchmarks just as R1-Distill-Qwen-7B's do, from around 13k tokens on AIME25 to 4k on MATH500. Denser adapters achieve slightly higher performance, and the remaining accuracy gap is comparable to the MLP fine-tuning skyline, suggesting it reflects training data differences rather than limitations of transcoder adapters. Despite sharing all non-MLP parameters with the target reasoning model, the hybrid baseline shows little sign of reasoning. The hybrid model exhibits slightly increased response lengths and a moderate increase in accuracy gains on select math benchmarks.
\section{Interpreting Transcoder Adapters}
\label{sec:features}

Having verified that transcoder adapters faithfully approximate fine-tuning differences, we next use study adapter features \footnote{The sparsely activating neurons in our transcoder adapters} and the computation graphs they form. Features are commonly studied via their (maximally) activating text examples and the tokens promoted or inhibited by their decoder directions \citep{dunefsky2024transcoders,ameisen2025circuit}. We collect max-activating and uniformly sampled examples for each feature over a held-out set of 5k Open Thoughts samples ($\sim$37M tokens). We use these activating examples for all subsequent analyses.
When using transcoder adapters to study model differences, each feature's computation is unique to the target model by construction. However, a feature's activating examples suggest what inputs trigger new computation---not whether the underlying representations are unique to the target model or already present in the base model.

\subsection{Automated Evaluations of Feature Interpretability}
\label{sec:feature_eval}

\begin{figure}[t]
    \centering
    \includegraphics[width=0.55\linewidth]{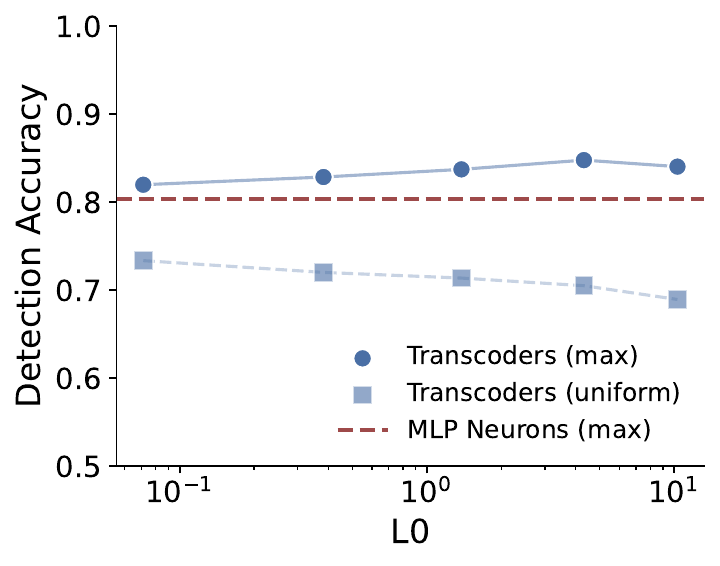}
    \caption{\textbf{Automated interpretability scores of transcoder adapters features and MLP neurons.} Max-activating examples of adapter features achieve slightly higher detection accuracy than neurons, while uniformly sampled activating examples score below neurons but well above random chance (0.50).}
    \label{fig:autointerp}
\end{figure}

We evaluate feature interpretability using the automated detection pipeline introduced by \citet{bills2023language}. This pipeline first uses a language model to generate a description of a feature activating text examples, then evaluates whether this description can be used to detect whether new text activates the feature. For adapter features, we compute detection scores using both max-activating and uniformly sampled activating text; for MLP neurons, we compute only max-activating scores. Additional details are in Appendix~\ref{sec:autointerp_details}. As shown in Figure~\ref{fig:autointerp}, max-activating examples of adapter features achieve detection scores of 0.82--0.84, slightly higher than MLP neurons at 0.80. Uniformly sampled activating examples score 0.69--0.73, lower than neurons but well above chance at 0.50.

\subsection{Transcoder Adapter Feature Classes}

\begin{figure*}[t!]
    \centering
    \begin{subfigure}[c]{0.3\textwidth}
        \centering
        \includegraphics[width=\linewidth, trim=0cm 0cm 0cm 0.3cm, clip]{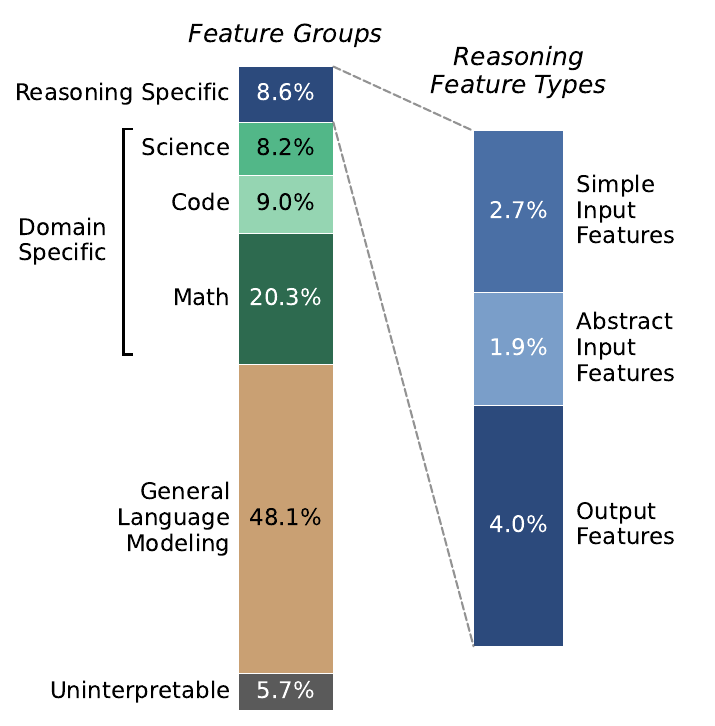}
        \caption{}
        \label{fig:qualitative_features_taxonomy}
    \end{subfigure}
    \hspace{0.4em}
    \begin{subfigure}[c]{0.66\textwidth}
        \centering
        \includegraphics[width=\linewidth, trim=1.5cm 18.3cm 1.5cm 0.2cm, clip]{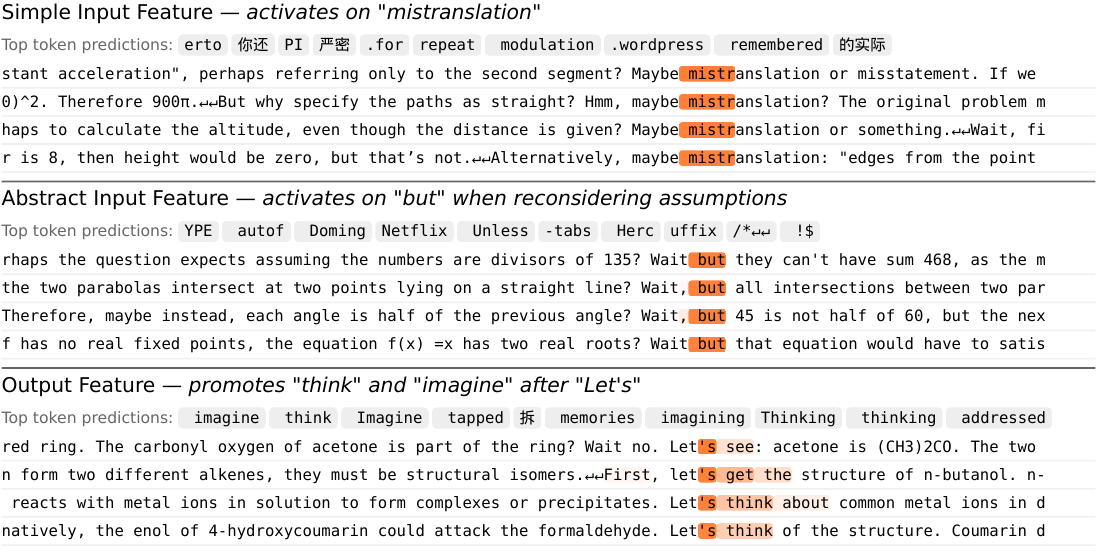}
        \caption{}
        \label{fig:qualitative_features_examples}
    \end{subfigure}
    \caption{\textbf{Transcoder adapter features.} \textbf{(a)} Approximate classification (via LLM judge) of transcoder adapter features into general language, domain-specific, and reasoning-related groups. Only a small fraction (8.6\%) are classified as reasoning-specific. We additionally categorize reasoning features by their function. \textbf{(b)} Select examples of reasoning features.\ificmlshowauthors{} All features can be browsed \href{https://transcoder-adapters.github.io/features/?feature=20476786}{here}.\fi}
    \label{fig:qualitative_features}
\end{figure*}

We next present several classes of adapter features, identified through manual inspection. We estimate class frequencies by classifying 7,000 features (250 per layer) from our $L_0$=1.4 adapter with an LLM judge. LLM judge details and 8 randomly sampled features from each class are shown in Appendix~\ref{sec:llm_judge} and  Appendix~\ref{sec:feature_dashboards} respectively. The examples in Figure~\ref{fig:qualitative_features_examples} are selected from these samples.

Around 48\% of features are indistinguishable from features one might find in a non-reasoning language model. These features activate on punctuation or generic terms, or promote common words. Another 37\% activate on technical content in mathematics, science, and code. Such domain-specific features have been observed in general language models \citep{templeton2024scaling}, but their high frequency among adapter features is notable. A smaller set of 8.6\% appear directly related to distinctive behaviors of reasoning models. We roughly categorize these as: \textit{output} features, which promote tokens like \textit{Wait}; \textit{simple input} features, which activate on specific reasoning words or phrases; and \textit{abstract input} features, which detect more abstract stages of reasoning, such as when revisiting assumptions. Select examples are shown in Figure~\ref{fig:qualitative_features_examples}. While feature counts based on activating text examples are only a rough proxy, we find it notable that so few adapter features appear related to reasoning, despite the adapters approximating the difference between a base and reasoning model. By this measure, more of the effect of reasoning training is due to increased domain knowledge than to reasoning-specific mechanisms.

\begin{figure}[!b]
    \centering
    \includegraphics[width=0.85\linewidth, trim=0.26cm 0 0 0, clip]{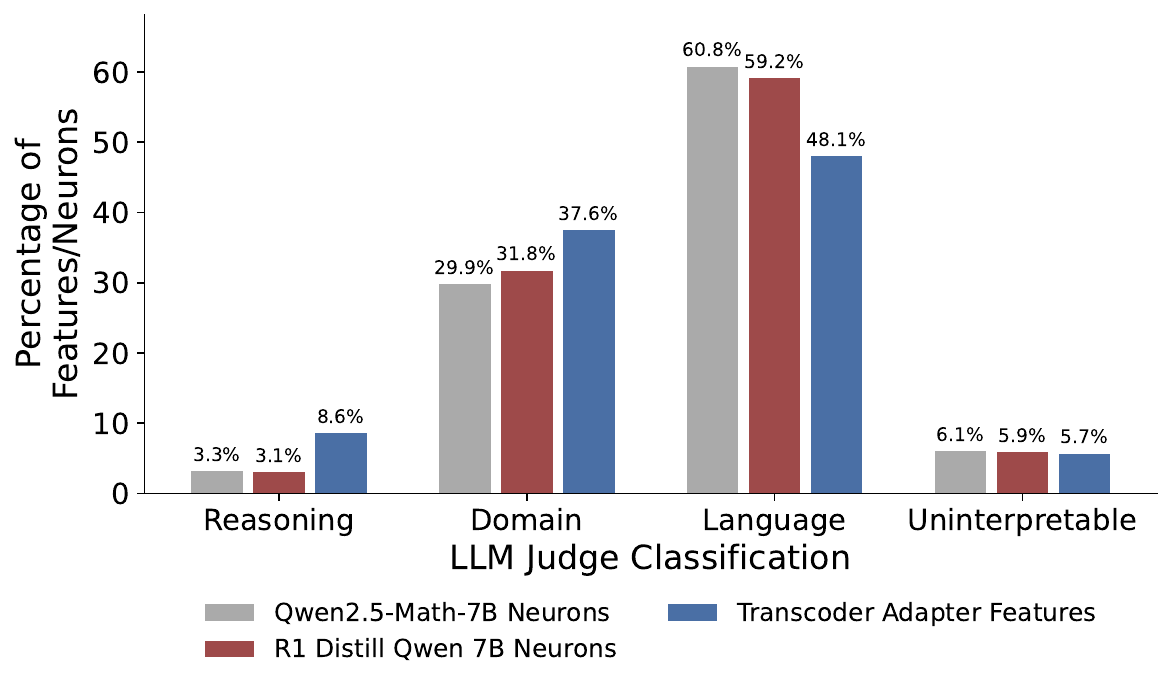}
    \caption{\textbf{Feature classification comparison.} Comparison of LLM judge classifications of transcoder adapter features with MLP neurons in the base and reasoning models. MLP neurons in the base and reasoning models have comparable class frequencies, suggesting reasoning fine-tuning has relatively small effects on overall model computation. More transcoder adapter features are classified as reasoning or domain-specific than MLP neurons in either model, suggesting adapters capture changes specific to reasoning fine-tuning}
    \label{fig:feature_taxonomy_comparison}
\end{figure}

As a sanity check, we classify MLP neurons in both the base and reasoning models using the same LLM judge (Figure~\ref{fig:feature_taxonomy_comparison}). MLP neurons in the two models have comparable class frequencies (~3\% reasoning-specific in both). As above, this is an imperfect proxy, but suggests that reasoning fine-tuning has relatively small effects on overall model computation. We do observe a slight increase in domain-specific neurons in the reasoning model, consistent with the high frequency of domain-specific adapter features. More adapter features are classified as reasoning-specific (~8\%) than neurons in either model, suggesting adapters capture changes specific to reasoning fine-tuning. 


\subsection{Attribution Graphs}
\label{sec:attribution_graphs}

\begin{figure*}[t!]
    \centering
    \begin{subfigure}[c]{0.375\linewidth}
        \centering
        \includegraphics[width=\linewidth]{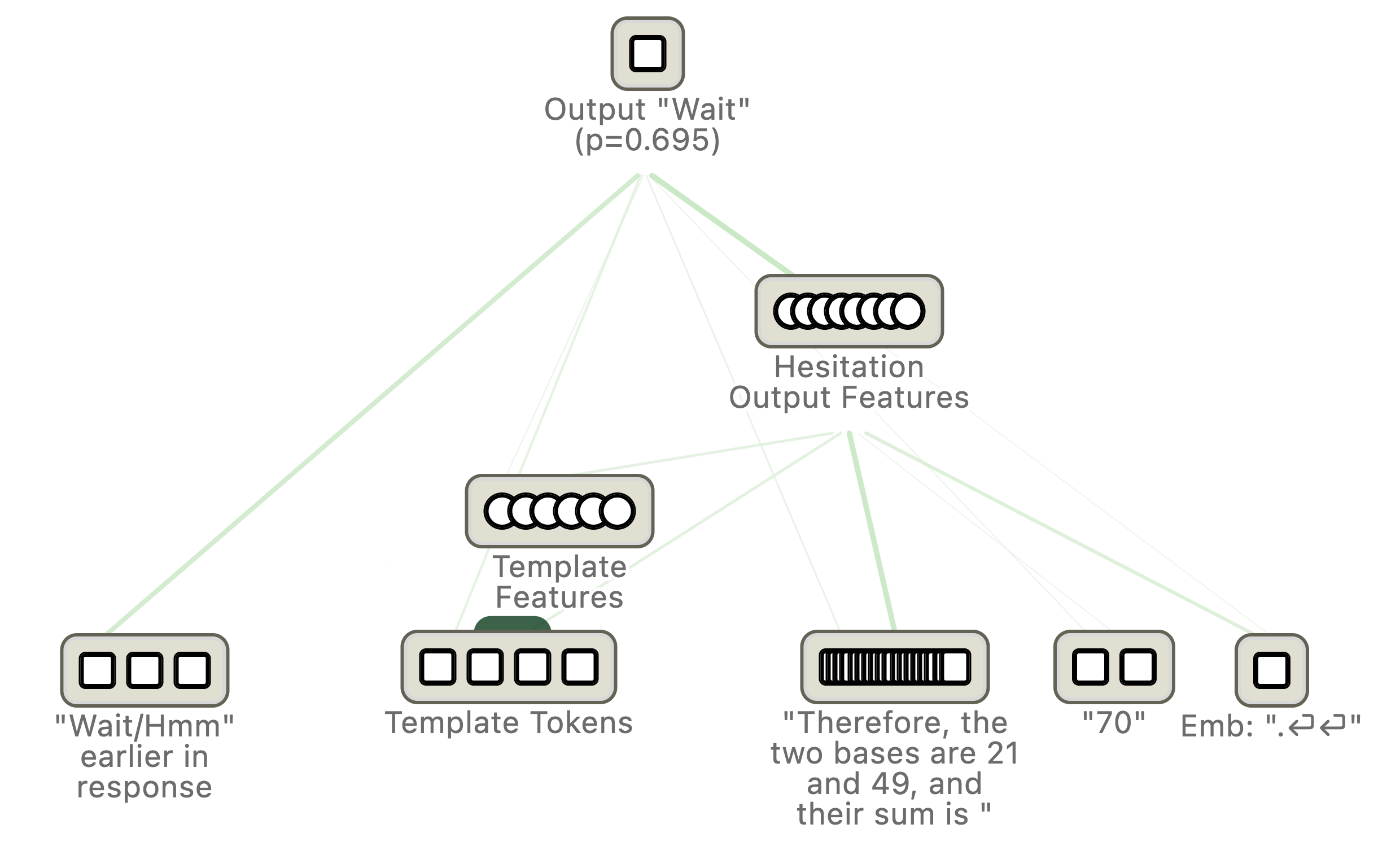}
        \caption{}
        \label{fig:attribution_graph_graph}
    \end{subfigure}
    \hspace{0.2cm}
    \begin{subfigure}[c]{0.575\linewidth}
        \centering
        \includegraphics[width=\linewidth, trim=1cm 21.2cm 3.5cm 0.2cm, clip]{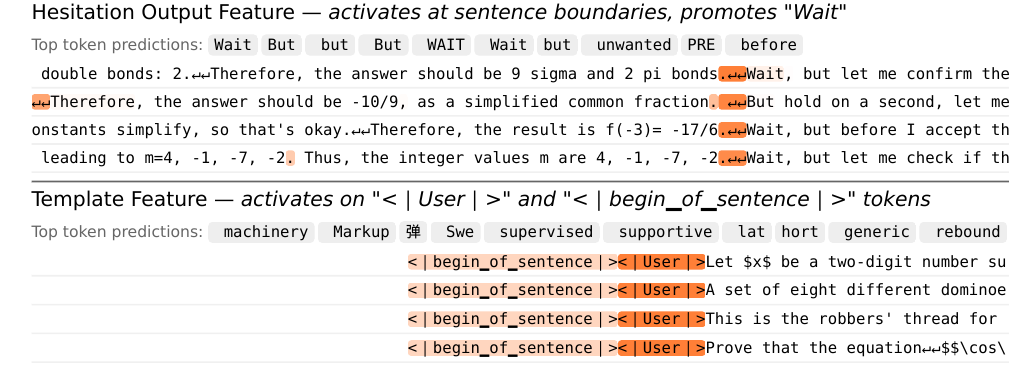}
        \caption{}
        \label{fig:attribution_graph_features}
    \end{subfigure}
    \caption{\textbf{Attribution graph for backtracking.} \textbf{(a)} Attribution graph for the \textit{Wait} prediction in ``\textit{Find the sum of all integer bases $b > 9$ for which $17_b$ is a divisor of $97_b$. \ldots\ Therefore, the two bases are 21 and 49, and their sum is 70. \textbf{Wait}}.'' The prediction primarily depends on two classes of adapter features: \textit{hesitation output features} promoting the token \textit{Wait} and \textit{template features} active on prompt formatting tokens. \textbf{(b)} Sample hesitation output features (top) and template features (bottom) with activating text.\ificmlshowauthors{} Full graph \href{https://transcoder-adapters.github.io/graphs/?slug=r1-qwen-7b-l0_1.4__wait_after_answer_main}{here}.\fi}
    \label{fig:attribution_graph_wait}
\end{figure*}

Unlike prior circuit analysis, which aims to give a complete account of the computation leading to an output \cite{lindsey2025biology,marks2025featurecircuits}, we build attribution graphs that capture the difference in computation between the base and target models. These graphs abstract away base model computation, viewing it simply as a medium through which nodes (token embeddings, transcoder adapter features, and final logits) can interact. 
We compute edge attribution using relevance propagation \citep{arora2025language,jafari2025relpfaithfulefficientcircuit}. 
Notably, attribution between nodes in our graph is propagated through base model MLP parameters, so edges capture not only direct effects through the residual stream but also indirect effects mediated by base model MLPs. We study attribution graphs using the implementation provided by \citet{circuit-tracer}. 

We present a case study of backtracking behavior, using our $L_0 = 1.4$ transcoder adapter to sample the response and construct the attribution graph\footnote{One could also sample from the target model and use a transcoder adapter to construct the attribution graph, though this would additionaly require error nodes \citep{lindsey2025biology} to account for reconstruction error between the adapter and the target model.}. When solving 2025 AIME I Problem 1, \textit{find the sum of all integer bases $b > 9$ for which $17_b$ is a divisor of $97_b$}, after $\sim$1100 tokens of reasoning, the model (correctly) deduces, \textit{Therefore, the two bases are 21 and 49, and their sum is 70.} The most likely next token, however, is \textit{Wait}. 
We construct an attribution graph to understand the causes of this predicted token (Figure~\ref{fig:attribution_graph_graph}).

In the attribution graph, the largest contributors to predicting \textit{Wait} are transcoder adapter features whose decoder directions promote backtracking tokens. We refer to these features as \textit{hesitation output features}. These features depend on very few other adapter features. We observe incoming edges from what we term \textit{template features}---features whose activating examples are exclusively on formatting tokens in the user or system prompt (Figure~\ref{fig:attribution_graph_features}). The remaining attribution to hesitation output features flows directly from token embeddings, suggesting that the features rely primarily on base model representations rather than on other adapter computation. This supports previous findings that reasoning fine-tuning repurposes pre-existing base model representations for backtracking \citep{ward2025reasoningfinetuningrepurposeslatentrepresentations}. Attribution also flows from earlier backtracking token embeddings to the \textit{Wait} logit, consistent with base model parameters performing induction.

It is worth dwelling on what is absent from this graph. We do not observe meaningful attribution from the embedding of the proposed answer (``70''), nor any adapter features performing verification or expressing uncertainty. We observe qualitatively similar graphs in other settings where the model says \textit{Wait}, including in the middle of reasoning and when seemingly confused by a typo in the prompt\ificmlshowauthors\footnote{Additional attribution graphs available on the project website.}\fi. This suggests that these two feature classes are broadly responsible for the model's hesitation behavior.

\section{Manipulating Transcoder Adapters}
\label{sec:interventions}

Our study of attribution graphs suggests that hesitation depends on two classes of features: hesitation output features promoting hesitation tokens and template features active at constant positions in the prompt (Section~\ref{sec:attribution_graphs}). We test this hypothesis through interventions on the $L_0 = 1.4$ transcoder adapter. We begin by systematically identifying all such features among the 229k transcoder adapter features using objective criteria:
\begin{itemize}
    \item \textbf{Hesitation output features:} features with one of four hesitation words (\textit{wait}, \textit{hmm}, \textit{but}, or \textit{alternatively}) among their top 10 promoted tokens (4,812 total).
    \item \textbf{Template features:} features with $\geq$80\% of max-activating examples at a consistent chat template position (811 total).\footnote{Positions checked: \texttt{<|begin\_of\_sentence|>}, \texttt{<|User|>}, \texttt{<|Assistant|>}, \texttt{<think>}, the newline following \texttt{<think>}, and the first content token after \texttt{<think>}.}
\end{itemize} For a given set of features, we conduct two interventions: removing them from the full transcoder adapter, and adding them to the hybrid model. In practice, both interventions produce a modified transcoder adapter: we zero out the encoder and decoder parameters corresponding to either the selected features or all other features. We can then sample from the resulting model and evaluate on benchmarks, measuring effects on model behavior over long rollouts rather than at single positions.

\begin{figure}[!t]
    \centering
    \includegraphics[width=0.9\linewidth]{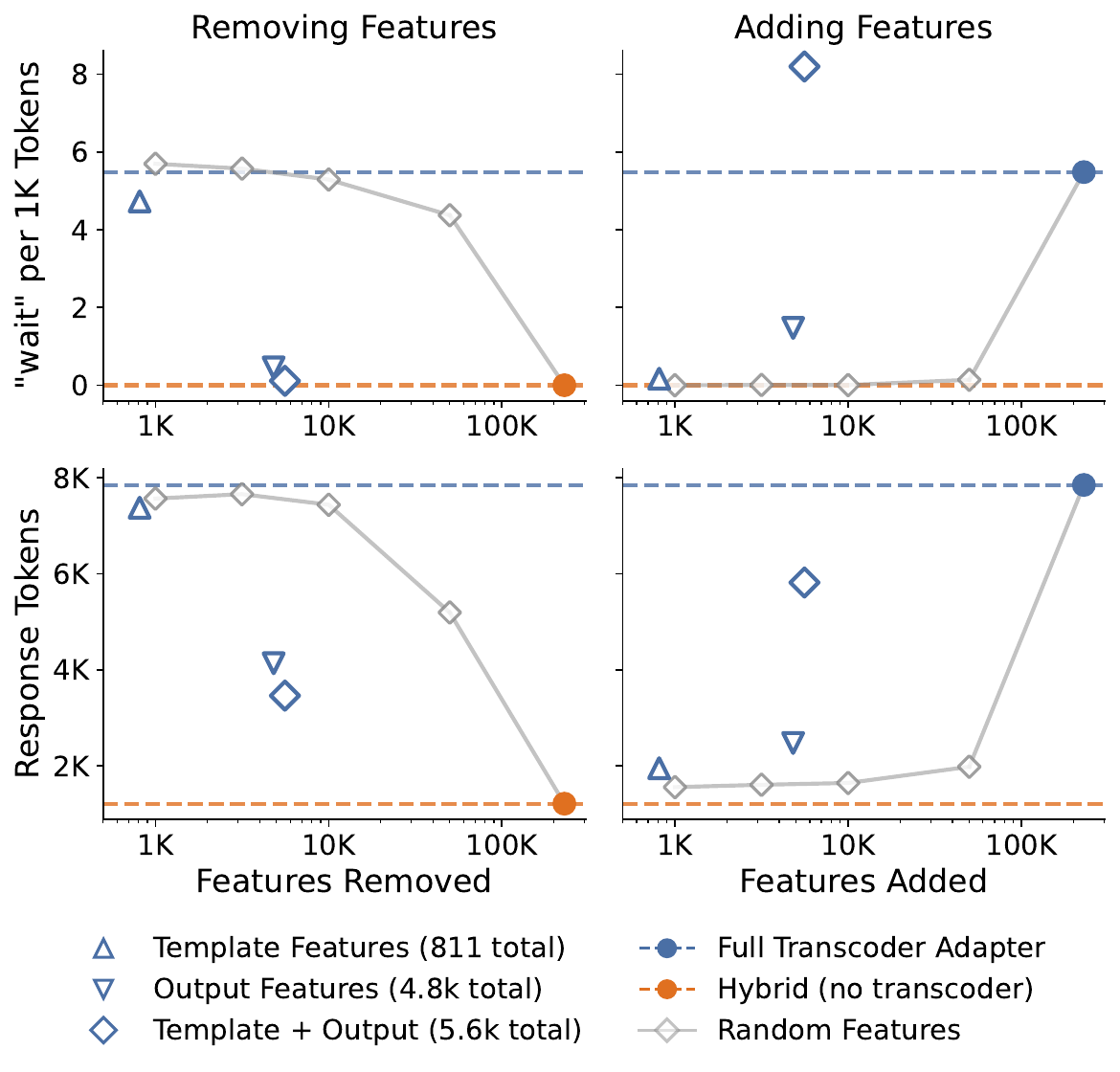}
    \caption{\textbf{Effects of removing or adding select adapter features.} \textbf{(Left)} Removing hesitation output features significantly reduces ``Wait'' frequency and response length. \textbf{(Right)} Adding both feature groups together produces much larger effects than either alone, suggesting they interact and together are sufficient to increase response length and ``Wait'' token frequency. Adding or removing random features (gray) produces much smaller effects.}
    \label{fig:backtracking_sufficiency}
\end{figure}

\subsection{Features Necessary and Sufficient for Hesitation}


As a proxy for hesitation, we measure \textit{wait} token frequency per 1,000 response tokens. Removing hesitation output features reduces \textit{wait} frequency by ~70\% (from 5.5 to 1.5); additionally removing template features reduces it further to 0.5. As the model generates fewer hesitation tokens, response length drops from 7.8k to 3.5k tokens when removing both feature groups (Figure~\ref{fig:backtracking_sufficiency}, left). Given that hesitation output features were selected precisely for promoting hesitation words, it is perhaps unsurprising that they are necessary for the model to say \textit{wait}. We next test whether these features are sufficient by adding them to the hybrid model, which has a near-zero rate of hesitation tokens and short responses (~1.2k tokens). Adding just template features or just hesitation output features produces slight increases (0.2 and 1.5, respectively), but adding both produces a much larger effect (8.2), supporting our observation that these feature classes interact (Figure~\ref{fig:backtracking_sufficiency}, right).\footnote{In fact, rate of \textit{wait} exceeds the full adapter's rate of 5.5. We find that adding an additional 5k features that activate before hesitation tokens returns the rate to full adapter levels without changing response length, suggesting that these additional features diversify continuation phrases.} When adding or ablating random features as a control, modifying even 50,000 random features produces smaller effects than our 5,623 hand-selected features.

\begin{figure}[tp]
    \centering
    \includegraphics[width=0.95\linewidth]{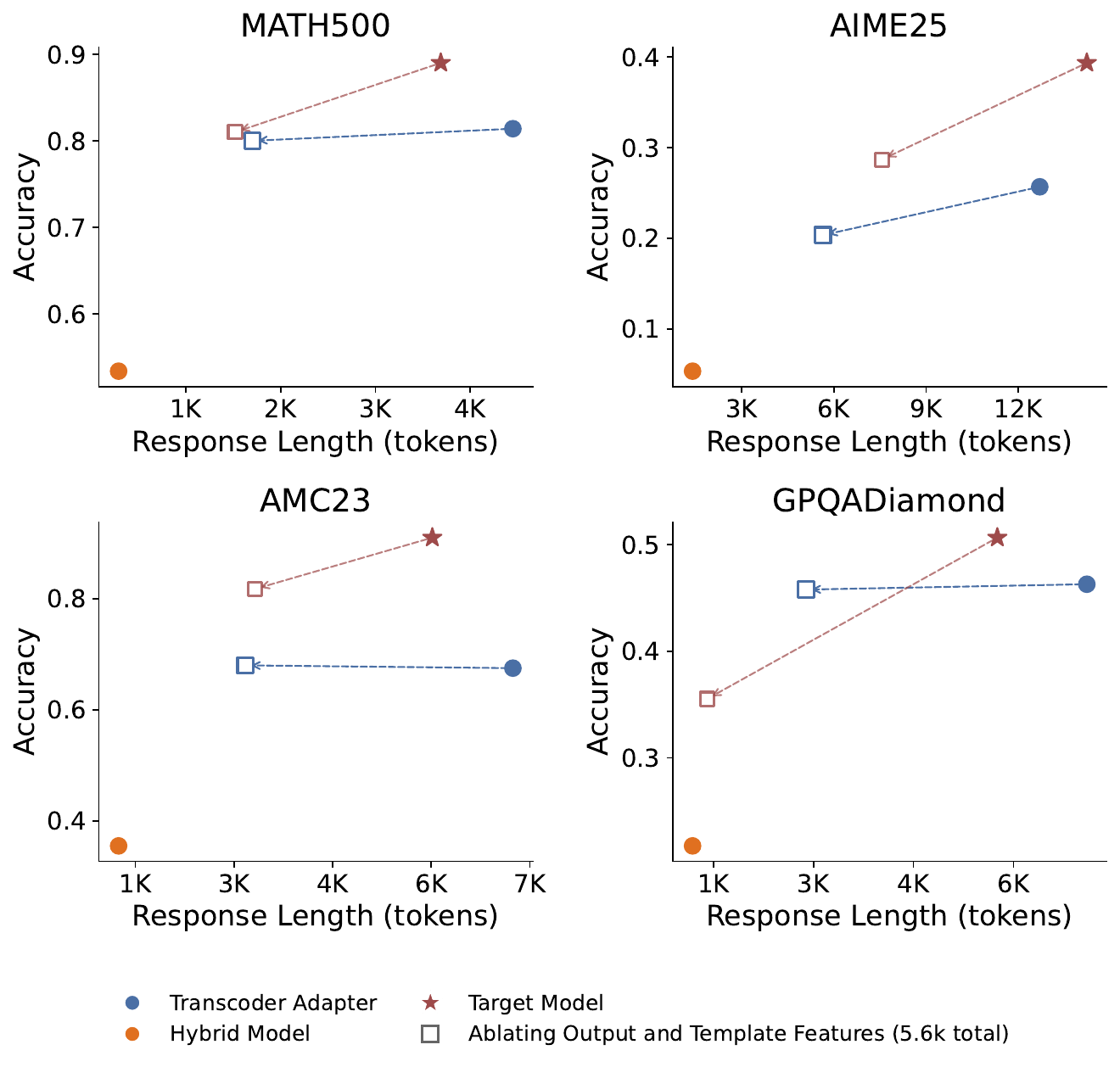}
    \caption{\textbf{Ablating hesitation features reduces response length while preserving accuracy.} Ablating template and hesitation output features from the transcoder adapter (blue) reduces response length without hurting accuracy on three of four benchmarks. Suppressing these features in the target model (red) also reduces response length, with a slightly larger accuracy drop.}
    \label{fig:backtracking_ablations}
\end{figure}

\subsection{Shortening Responses with Feature Ablation}

Having traced hesitation tokens and increased response length to a set of 5.6k hesitation output and template features, we test whether ablating these features can shorten responses without harming accuracy. We ablate the features from the transcoder adapter as in the previous experiments. Across benchmarks, response length drops by more than 50\% (e.g., 7.2k to 3.2k tokens on AMC23), while accuracy is unchanged on MATH500, AMC23, and GPQA Diamond. On AIME25, accuracy decreases from 26\% to 20\%. This is consistent with prior work showing it is possible to reduce reasoning-model verbosity, via additional training \citep{liu2025dlerdoinglengthpenalty} or inference-time interventions \citep{wang2025waitdontneedwait, huang2025mitigatingoverthinkinglargereasoning}. Though less performant, our intervention emerges naturally from decomposing the fine-tuning difference, suggesting that a partial separation between verbosity and accuracy arises naturally from reasoning training.

We next test whether this intervention transfers to the target model. In the adapter setting, we ablate features by zeroing their encoder and decoder parameters. We observe that ablating transcoder adapter features is equivalent to adding a new feature with the same encoder and a negated decoder. We thus intervene on the target model by first constructing an adapter where each feature we wish to suppress is negated\footnote{Scaling by $-1$ led to incoherent outputs; we use $-0.5$. We hypothesize this is because of distribution shift in encoder inputs.}, then evaluating the target model with this negated adapter added. Figure~\ref{fig:backtracking_ablations} (red) shows that suppressing these adapter features in the target reasoning model also reduces response length, with only a slight decrease in accuracy. We expect some decrease in performance given this cruder intervention. However, accuracy remains high enough to confirm that our findings transfer to the target model without severely damaging its reasoning capabilities.

\section{Conclusion}
\label{sec:conclusion}


We introduce transcoder adapters, a method for learning sparse, interpretable approximations of the changes in MLP computation learned during fine-tuning. Applying them to Qwen2.5-Math-7B and DeepSeek-R1-Distill-Qwen-7B, we verify that adapters faithfully capture the target model's outputs, internal states, and qualitative behavior while being remarkably sparse and interpretable, achieving higher automated detection scores than MLP neurons. We use adapters to study reasoning, finding that only a small fraction of adapter features have activating examples related to reasoning behaviors, suggesting much of the change may stem from domain knowledge. We rigorously characterize hesitation behavior, finding it is governed by surprisingly simple and interpretable mechanisms. 

\textbf{Limitations and Future Work.} While transcoder adapters give a full account of changes in MLP computation, changes to non-MLP parameters—most notably attention, but also embeddings—remain unexplained. In this work, we rigorously confirm via our hybrid baseline that changes to embedding and attention parameters alone are insufficient to explain the reasoning behavior learned by the model. While recent work has begun decomposing attention using sparse methods \citep{he2025understandingnatureattentionlowrank}, extending this to study differences between models remains an open question. Our work also focuses on deeply studying a single base and reasoning model pair: Qwen2.5-Math-7B and DeepSeek-R1-Distill-Qwen-7B. Notably, this model is relatively small and trained via distillation on transcripts from a larger reasoning model. Transcoder adapters offer an exciting tool to study fine-tuning more broadly—pre-training stages, human-assistant post-training, or effects like emergent misalignment \citep{Betley_2026}. We expect transcoder adapter training to benefit from sparse dictionary learning advances: JumpReLU nonlinearities \cite{rajamanoharan2024jumpingaheadimprovingreconstruction} and auxiliary losses to prevent dead latents \cite{gao2024scalingevaluating} would likely improve quality directly, while approaches incorporating cross-layer representations \cite{ameisen2025circuit} or Matryoshka-style objectives \cite{bussmann2025learningmultilevelfeaturesmatryoshka} would be exciting though less straightforward to incorporate.

\section*{Impact Statement}
This paper presents work that advances language model interpretability. Improved understanding of model internals can benefit AI safety and broader machine learning research. More generally, there are many potential societal consequences of advancing the field of machine learning, none of which we feel must be specifically highlighted here.

\ificmlshowauthors
\section*{Acknowledgements}
We thank Percy Liang for helpful discussions and feedback during the early stages of this project. This research is supported in part by a grant from Open Philanthropy.
\fi

\clearpage
\bibliography{references}
\bibliographystyle{icml2026}

\newpage
\appendix
\onecolumn
\section{Training Details}
\label{sec:training_details}

\paragraph{Optimization.} We train transcoder adapters using the Adam optimizer with default hyperparameters, a learning rate of $8 \times 10^{-4}$, and a batch size of 1. Although we use a batch size of 1, OpenThoughts3 samples average approximately 7,500 tokens, so each gradient step computes losses over all token positions in the sample. We apply a linear learning rate warmup for the first 5\% of training followed by cosine decay. Training is conducted in bfloat16 precision.

\paragraph{Loss computation.} Computing the bridging loss at every layer is expensive, as it requires forward passes through the remaining layers of both models. To reduce this cost, we estimate the bridging losses at each step by uniformly sampling a single cutoff layer $k \sim \text{Uniform}\{1, \ldots, L\}$ and computing the forward and backward bridging losses at layer $k$ only. We weight all four reconstruction losses (output KL, forward and backward bridging KL, and NMSE) equally with coefficient 1.

\paragraph{Sparsity.} We train adapters at five L1 sparsity coefficients: $\{0.01, 0.003, 0.001, 0.0003, 0.0001\}$.

\paragraph{Dataset.} OpenThoughts3 \citep{guha2025openthoughtsdatarecipesreasoning} is a curated dataset of reasoning responses from QwQ-32B \citep{qwq32b}, with sequences up to 16,384 tokens. We construct separate training (50,000 samples) and validation (5,000 samples) sets, both filtered for sequences under 10,000 tokens, with domain proportions matching the full dataset (71\% math, 21\% code, 8\% science). The validation set is used to collect feature-activating examples for downstream analysis.

\paragraph{MLP fine-tuning skyline.} For the MLP fine-tuning skyline (Section~\ref{sec:evaluation}), we use the Adam optimizer with the same default hyperparameters and learning rate schedule as adapter training. We sweep over learning rates in $\{3 \times 10^{-3}, 1 \times 10^{-3}, 3 \times 10^{-4}, 1 \times 10^{-4}\}$, finding $1 \times 10^{-4}$ to be optimal.

\section{Hybrid Baseline Variants}
\label{sec:hybrid_appendix}

Our transcoder adapter methodology is limited to decomposing differences in MLP computation. For our reasoning case study, we must confirm that MLP computation is critical—that attention changes alone do not elicit reasoning. This motivates the hybrid baseline.

To ensure the hybrid baseline's limited performance is not simply due to combining parameters naively, we evaluate several variants (Table~\ref{tab:hybrid_variants}). We test few-shot prompting using the same prompts as base model evaluation, at both temperature 0 and 0.7. We also fine-tune only the RMSNorm parameters on 1M tokens (Hybrid + RMS Refit). No variant significantly increases benchmark performance or response length in ways indicative of reasoning behavior.

\begin{table}[h]
\centering
\caption{\textbf{Hybrid baseline variants.} Accuracy (\%) and response length (tokens) across benchmarks (AIME 2025, AMC 2023, MATH-500, GPQA Diamond). Few-shot prompting and RMSNorm refitting fail to elicit reasoning behavior from hybrid models.}
\label{tab:hybrid_variants}
\begin{tabular}{lcccc|c}
\toprule
& AIME 25 & AMC 23 & MATH-500 & GPQA Diamond & Avg \\
\midrule
\multicolumn{6}{l}{\textit{Accuracy (\%)}} \\
Hybrid (zero-shot) & 5.0 & 35.5 & 56.0 & 21.2 & 29.4 \\
Hybrid + few-shot ($t$=0) & 0.0 & 7.5 & 32.2 & 19.2 & 14.7 \\
Hybrid + few-shot ($t$=0.7) & 0.0 & 12.5 & 29.4 & 23.1 & 16.3 \\
Hybrid + RMS Refit & 6.7 & 34.5 & 53.8 & 24.6 & 29.9 \\
\midrule
\multicolumn{6}{l}{\textit{Response Length (tokens)}} \\
Hybrid (zero-shot) & 1677 & 1178 & 999 & 1094 & 1237 \\
Hybrid + few-shot ($t$=0) & 388 & 431 & 513 & 218 & 387 \\
Hybrid + few-shot ($t$=0.7) & 380 & 514 & 491 & 143 & 382 \\
Hybrid + RMS Refit & 1549 & 1355 & 898 & 1154 & 1239 \\
\bottomrule
\end{tabular}
\end{table}

\section{Internal Faithfulness (Extended)}
\label{sec:faithfulness_appendix}

\begin{figure}[t]
    \centering
    \includegraphics[width=\linewidth]{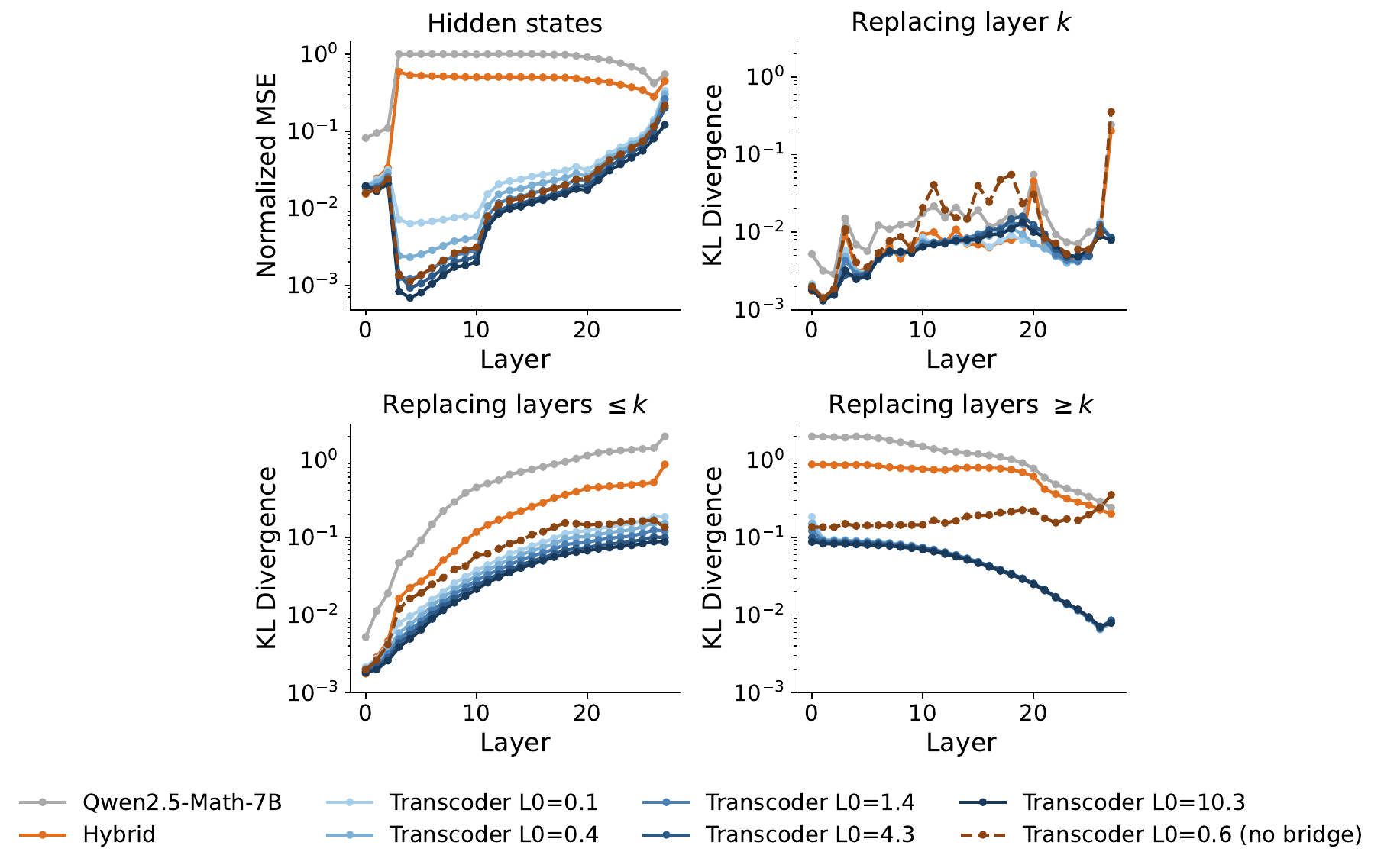}
    \caption{\textbf{Internal faithfulness metrics (extended).} Extended version of Figure~\ref{fig:internal_metrics} with all sparsity levels and a bridging loss ablation (dashed). All main adapters substantially outperform baselines. Ablating the bridging loss preserves low NMSE but increases KL divergence across replacement interventions, most notably when replacing the final $k$ layers with transcoder outputs.}
    \label{fig:internal_metrics_appendix}
\end{figure}

Figure~\ref{fig:internal_metrics_appendix} shows an extended version of the internal faithfulness metrics with all transcoder adapters at different sparsities, plus an ablation where we remove the bridging loss. Across sparsity levels, all main adapters substantially outperform baselines, with denser models achieving slightly lower error as expected. When ablating the bridging loss, raw hidden state reconstruction (NMSE) remains very low; however, KL divergence across various replacement interventions increases. This effect is most pronounced when replacing the final $k$ layers of the target model with transcoder layers, confirming that the bridging loss helps adapter outputs integrate with downstream computation.

\section{Evaluation Details}
\label{sec:eval_details}

We use Evalchemy \citep{evalchemy2025} for all evaluations. For the base model (Qwen2.5-Math-7B), we evaluate using few-shot prompts at temperature 0. For all other models—including the target model, transcoder adapters, and hybrid baselines—we use typical reasoning model configurations: temperature 0.7, 32,768 max sampled tokens, and top\_p=1.0. Table~\ref{tab:eval_samples} summarizes the number of questions and repetitions per benchmark.

The models we study occasionally produce degenerate text, especially hybrid models and sparse transcoder adapters. We detect degeneration using a sliding window of 200 tokens. For each window, we compute the coverage of the most frequent n-gram for $n \in \{1, 2, 3, 4, 5\}$, defined as $(\text{count} \times n) / \text{window size}$. A window is marked degenerate if any n-gram exceeds 25\% coverage, meaning roughly a quarter of the window is dominated by a single repeated pattern. If degeneration is detected, we truncate at the left edge of the earliest degenerate window. We still attempt to parse an answer from the truncated response; response length is measured up to truncation. This reflects a realistic deployment setting where models are served with additional early stopping criteria, such as lightweight degeneration classifiers.

Table~\ref{tab:degeneration_rates} reports truncation rates across models. As transcoder adapters become sparser, truncation rates increase; the hybrid baseline has the highest truncation rate. Surprisingly, despite substantial variation in truncation rates across adapters, differences in benchmark accuracy are much smaller. We hypothesize that degeneration is more likely on harder questions that neither model would answer correctly.
    
\begin{table}[h]
\centering
\caption{\textbf{Evaluation sample counts.} Number of questions and repetitions per benchmark.}
\label{tab:eval_samples}
\begin{tabular}{lcccc}
\toprule
& AIME 25 & AMC 23 & MATH-500 & GPQA Diamond \\
\midrule
Questions & 30 & 40 & 500 & 198 \\
Repetitions & 10 & 10 & 1 & 3 \\
\bottomrule
\end{tabular}
\end{table}

\begin{table}[h]
\centering
\caption{\textbf{Degeneration rates.} Percentage of responses truncated due to degenerate text detection.}
\label{tab:degeneration_rates}
\begin{tabular}{lcccc|c}
\toprule
& AIME 25 & AMC 23 & MATH-500 & GPQA Diamond & Avg \\
\midrule
DeepSeek-R1-Distill (target) & 0.0\% & 0.8\% & 0.2\% & 0.0\% & 0.2\% \\
Dense Transcoder & 3.0\% & 1.2\% & 1.8\% & 0.3\% & 1.6\% \\
MLP Fine-tune & 3.0\% & 0.8\% & 0.4\% & 0.5\% & 1.2\% \\
Transcoder $L_0$=10.3 & 6.0\% & 3.2\% & 2.8\% & 1.3\% & 3.3\% \\
Transcoder $L_0$=4.3 & 13.0\% & 2.8\% & 2.0\% & 1.5\% & 4.8\% \\
Transcoder $L_0$=1.4 & 17.7\% & 9.2\% & 2.4\% & 4.0\% & 8.3\% \\
Transcoder $L_0$=0.4 & 27.0\% & 10.0\% & 5.0\% & 6.2\% & 12.1\% \\
Transcoder $L_0$=0.1 & 26.7\% & 11.5\% & 3.6\% & 6.2\% & 12.0\% \\
Hybrid (zero-shot) & 44.3\% & 31.0\% & 21.8\% & 29.3\% & 31.6\% \\
\bottomrule
\end{tabular}
\end{table}

\section{LLM Judge Details}
\label{sec:llm_judge}

We use GPT-5-mini \citep{singh2025openaigpt5card} as an LLM judge to classify transcoder adapter features into interpretable categories. The judge is presented with up to 10 max-activating text examples for each feature, along with the top logits promoted by the feature's decoder direction. The system and user prompts are reproduced below.

\begin{lstlisting}[title={\small\textbf{System Prompt}}]
You are a meticulous AI researcher analyzing neurons (features) in a sparse autoencoder trained on a mathematical reasoning model (DeepSeek-R1-Distill).

Your task is to classify features based on their activation patterns (where they fire) and output behavior (what tokens they promote).

## Level 1: Feature Domain

### "language" - General Language/Text Modeling
Features related to general language patterns that make text flow, not specific to math/code or reasoning model behaviors.

Key distinction from "reasoning": If a feature fires on generic language but ONLY in reasoning-model-specific contexts (e.g., triggers uncertainty, self-correction), classify it as "reasoning" not "language".

Examples of language features:
- Punctuation (commas, periods, quotes)
- Conjunctions (and, but, or, because)
- Articles and determiners (a, an, the)
- Pronouns and basic syntax
- Generic formatting (indentation, spacing)
- Connector/flow words: "therefore", "hence", "thus", "so", "then", "next"
- Standard prose transitions that make language flow smoothly

### "domain" - Math/Science/Code Technical Knowledge
Features encoding domain-specific technical vocabulary, notation, or patterns. This is about CONTENT knowledge, not language flow.

If you classify as "domain", also specify the domain_type:
- "math": Mathematical notation, equations, variables, math terms (sum, integral, =, "derivative", "polynomial", "let x be")
- "science": Scientific formulas, chemistry, physics terminology (chemical formulas, physical constants, scientific notation)
- "code": Programming syntax, keywords, operators ("def", "return", "if", brackets, indentation patterns)

Examples:
- Mathematical notation (sum, integral, forall, exists, =, +, numbers, variables) -> math
- Code syntax (keywords like "def", "return", "if", operators, brackets) -> code
- Technical math terms ("recursion", "derivative", "polynomial", "function", "matrix") -> math
- Scientific formulas and notation -> science
- Domain-specific structural patterns (equation layout, code blocks, LaTeX) -> math or code
- Math setup phrases: "let x be", "suppose", "given that", "define" -> math

### "reasoning" - Reasoning Model Behaviors
Features related to the unique behaviors of REASONING MODELS (like DeepSeek-R1), NOT general problem-solving.

KEY DISTINCTION:
- "language": Connector words and flow language ("therefore", "hence", "so", "then") - these make text flow but aren't unique to reasoning models.
- "domain": Technical math/code vocabulary and notation ("function", "derivative", "let x be") - this is content knowledge.
- "reasoning": Behaviors UNIQUE to reasoning models - the verbose, self-reflective, uncertainty-expressing style from RL/distillation training.

IMPORTANT: All examples are collected from reasoning traces, so co-occurrence is NOT enough. The feature must capture something unique to the REASONING MODEL STYLE:
1. Explicit uncertainty/hedging ("Hmm", "Wait", "I think", "maybe", "actually")
2. Self-correction and backtracking ("No, that's wrong", "Let me reconsider", "I made an error")
3. Metacognitive commentary ("Let me think about this", "I need to be careful here")
4. Verification behaviors ("Let me check", "Does this make sense?", "Sanity check")
5. The characteristic "think out loud" verbosity of reasoning models

NOT reasoning:
- Connector words ("therefore", "hence", "so", "thus") -> classify as "language"
- Technical terms ("the function", "derivative", "let x be") -> classify as "domain"

Examples of reasoning features:
- Uncertainty: "hmm", "wait", "actually", "I'm not sure"
- Self-correction: "no wait", "that's wrong", "let me reconsider"
- Metacognition: "I need to think about", "this is tricky"
- Verification: "let me verify", "checking my work", "does this make sense"
- Features that PROMOTE these reasoning-model-specific tokens

### "uninterpretable" - No Clear Pattern
The feature's firing pattern or role is unclear, even if the examples share surface-level similarities.

Signs to classify as uninterpretable:
- Examples share a theme (e.g., all math text) but you can't identify WHAT specifically triggers activation
- The highlighted token varies without a clear unifying pattern
- Top logits don't relate coherently to the activation pattern

As a very rough guide, prior work on SAE interpretability finds that typically 10-30% of features are uninterpretable. Do not anchor to this number, but don't force patterns that aren't there.

## Level 2: Mechanism (ONLY for "reasoning" features)

If you classified the feature as "reasoning", also classify its mechanism:

### "output" - Output Feature
The defining characteristic is WHAT it promotes. Has a clear output pattern (promotes specific tokens like "Wait", "Hmm", hesitation words). The input may vary but often has some pattern too.

How to identify:
- TOP LOGITS show it consistently promotes specific reasoning-related tokens
- The main story is "this feature promotes X" (input context is secondary)
- Most reasoning features that promote uncertainty/hesitation tokens fall here

Example: Promotes "Wait", "Hmm", "Hold on" - fires at various transition points but the key behavior is promoting these tokens.

### "input_simple" - Simple Input Feature
The defining characteristic is a simple TOKEN-LEVEL input pattern. Fires on specific tokens regardless of surrounding context.

How to identify:
- Fires on the SAME or SIMILAR tokens across examples (e.g., "wait", "Wait", "waiting")
- The token alone determines whether it fires - context doesn't matter
- Output may or may not be clear

Example: Fires on "actually" in any context within reasoning text.

### "input_abstract" - Abstract Input Feature
The defining characteristic is a CONTEXT-DEPENDENT input pattern. The same token might fire in some contexts but not others - broader context determines firing.

How to identify:
- Varied tokens across examples, but a consistent CONTEXTUAL theme
- The token alone does NOT determine firing - context matters
- Examples: "doesn't" only at logical contradictions, various planning phrases, transition points

Example: Fires on "doesn't" but ONLY when arriving at a logical contradiction (not every "doesn't").
Example: Fires on various phrases related to planning/strategizing (unified by concept, not token).

## Output Format

```json
{
    "reasoning": "Your step-by-step analysis...",
    "category": "language" | "domain" | "reasoning" | "uninterpretable",
    "confidence": "high" | "medium" | "low",
    "category_description": "Brief explanation of why this category",

    // ONLY include if category is "domain":
    "domain_type": "math" | "science" | "code",

    // ONLY include if category is "reasoning":
    "mechanism": "output" | "input_simple" | "input_abstract",
    "mechanism_description": "Explanation of the mechanism",
    "input_pattern": "What triggers it (for input_simple or input_abstract)",
    "output_pattern": "What it promotes (especially for output mechanism)"
}
```
\end{lstlisting}

\begin{lstlisting}[title={\small\textbf{User Prompt}}]
Analyze Feature L{layer}F{feature}:

## Top Logits (tokens this feature promotes when active):
{top_logits}

## Activating Examples (<<<token>>> marks where feature activates):
{examples}

First, classify this feature's domain (language, domain, reasoning,
or uninterpretable). If it's a reasoning feature, also classify its
mechanism (output, input_simple, or input_abstract).
\end{lstlisting}

%
%

\section{Automated Interpretability Details}
\label{sec:autointerp_details}

We evaluate feature interpretability using the automated detection pipeline introduced by \citet{bills2023language} and subsequently widely adopted for learned sparse dictionary features \citep{paulo2025automaticallyinterpretingmillionsfeatures,karvonen2025saebenchcomprehensivebenchmarksparse}. A first LLM call generates a natural language description from a feature's max-activating examples. A second LLM call presents the description alongside a mixture of positive samples that activate the feature and negative samples that do not, and the feature's detection score is the accuracy with which the LLM identifies which samples activate the feature.

We filter for features with activation frequency at least $6 \times 10^{-7}$ and sample 100 features from each layer. We use GPT-4o-mini as both the description generator and the detector.

For the description stage, we present up to 10 max-activating exemplars, each consisting of up to 71 tokens (50 preceding and 20 subsequent tokens surrounding the max-activating token).

For the detection stage, the model is presented with a shuffled sequence of 10 texts: 5 known to activate the feature and 5 randomly sampled from Open Thoughts. The detection score is the accuracy with which the model identifies the activating samples.

The description generation and detection prompts are reproduced below.

\begin{lstlisting}[title={\small\textbf{Description Generation: System Prompt}}]
You are a meticulous AI researcher investigating a specific neuron inside a language model. Your task is to describe what causes the neuron to activate.

You will receive text excerpts where the neuron activates. The activating token is marked with <<<token>>>.

Important notes:
- The <<<>>> markers are ONLY for highlighting which token activates - do NOT include <<<>>> in your description
- All examples are from mathematical reasoning contexts, so "math" or "reasoning" alone is NOT a useful description
- Neuron activations can only depend on the marked token and tokens BEFORE it (not after)
- Describe BOTH the general context AND the specific token/word/phrase that activates
- Be extremely specific: look for specific tokens, characters, syntactic positions, semantic patterns, or reasoning steps
- Descriptions should be 10-15 words, no need for complete sentences

Respond in JSON format.
\end{lstlisting}

\begin{lstlisting}[title={\small\textbf{Description Generation: User Prompt}}]
Neuron L{layer}F{feature}:

{examples}

Respond with:
{
    "reasoning": "brief analysis of patterns you see",
    "description": "concise description (10-15 words)"
}
\end{lstlisting}

\begin{lstlisting}[title={\small\textbf{Detection: System Prompt}}]
You are evaluating whether a neuron description accurately predicts neuron activations.

You will be given:
1. A description of what causes a neuron to activate
2. 10 text snippets (exactly 5 activate the neuron, 5 do not)

For each snippet, predict whether the neuron activates based ONLY on the description.

Respond with a JSON object mapping snippet numbers to predictions:
{"1": true, "2": false, "3": true, ...}
\end{lstlisting}

\begin{lstlisting}[title={\small\textbf{Detection: User Prompt}}]
Neuron description: {description}

Snippets (5 activate, 5 don't):
{snippets}

For each snippet, does the neuron activate? Respond with JSON: {"1": true/false, "2": true/false, ...}
\end{lstlisting}

\section{Feature Activation Overlap}
\label{sec:activation_overlap}

\begin{figure}[t]
    \centering
    \includegraphics[width=0.55\linewidth]{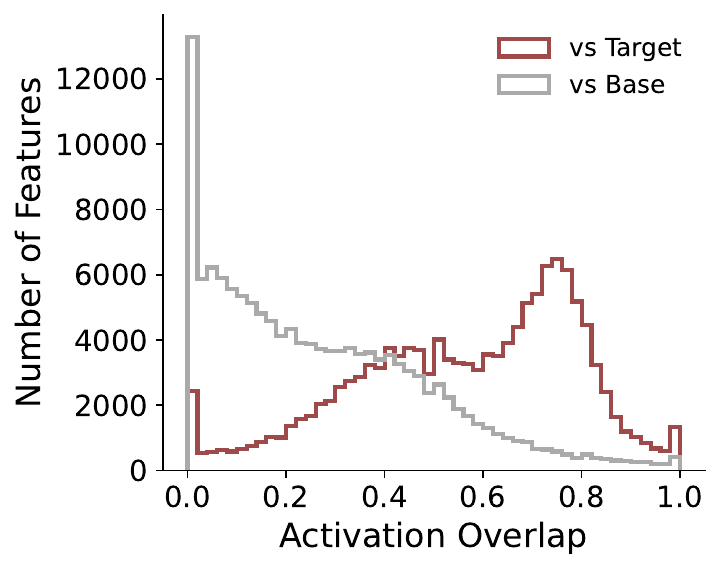}
    \caption{\textbf{Feature activation overlap.} Agreement rates are higher when recomputing adapter feature activations using target model hidden states than base model hidden states.}
    \label{fig:feature_alignment}
\end{figure}

For each feature, we measure how consistently it activates when computed from adapter versus target or base model hidden states. On a dataset, we first identify the inputs where the feature fires under the adapter—say $x$\% of inputs. We then run the same dataset through the target model (or base model), project hidden states onto the feature's encoder direction, and take the top $x$\% by activation magnitude. This is equivalent to refitting the encoder bias to match the original activation frequency; without this, agreement rates would be trivially inflated. The agreement rate for each feature is the overlap between its top-activating inputs under the adapter versus under each model. Figure~\ref{fig:feature_alignment} shows the distribution of agreement rates across all features. Agreement rates are substantially higher when recomputing feature activations using target model hidden states than base model hidden states, confirming that adapter features capture computation specific to the target model.

\section{Feature Dashboards}
\label{sec:feature_dashboards}

We present randomly sampled feature dashboards from each LLM judge category (see Section~\ref{sec:llm_judge} for classification details). Each dashboard shows the feature's top logits and example activating contexts.

\begin{figure*}[h!]
    \centering
    \includegraphics[width=0.9\linewidth, trim=1.4cm 6.2cm 1.4cm 0cm, clip]{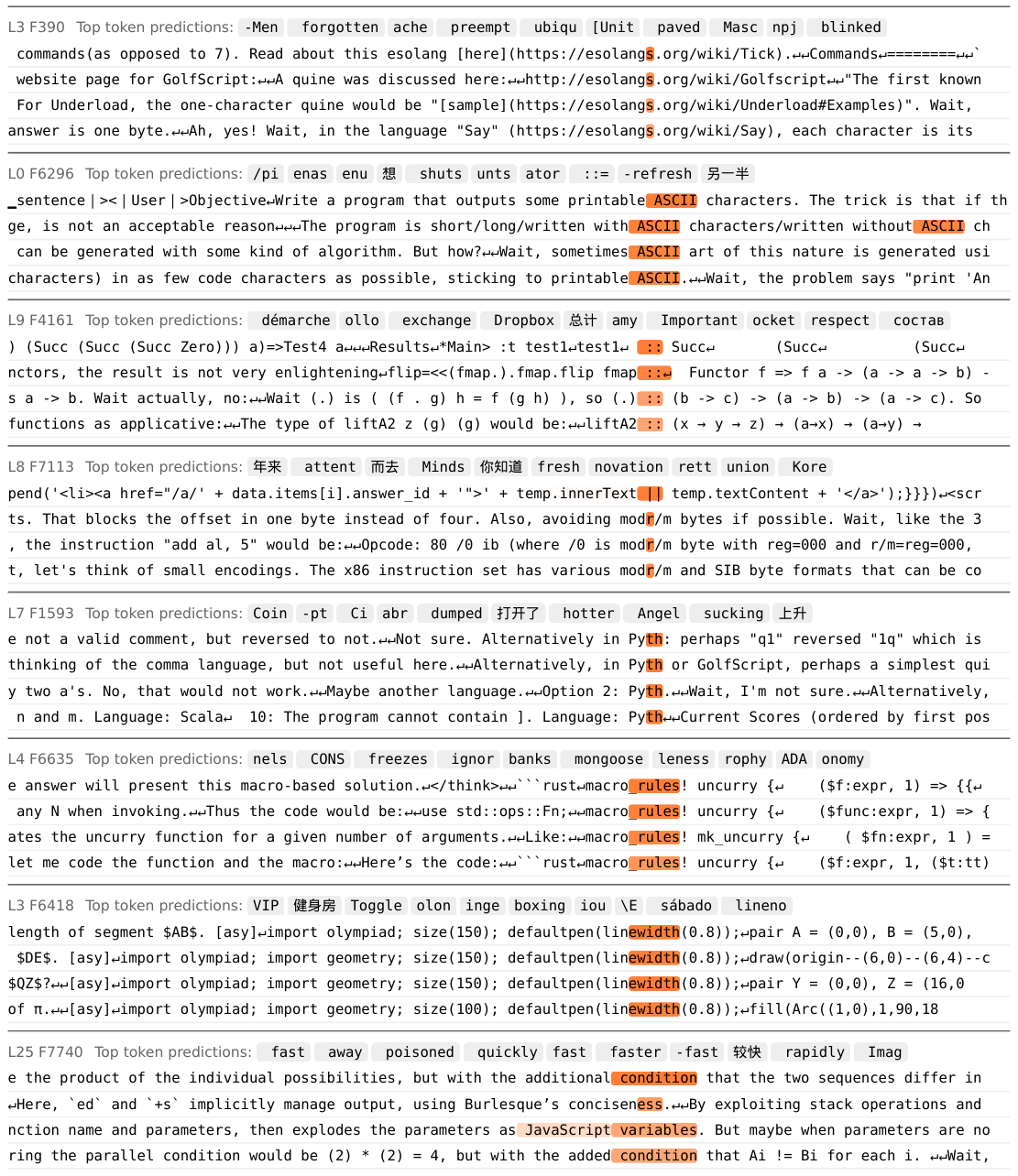}
    \caption{\textbf{Domain-specific features: code.} Randomly sampled transcoder adapter features classified by the LLM judge as domain-specific to code. For each feature, we show four max-activating dataset examples alongside the tokens most promoted by the feature's decoder direction.}
    \label{fig:dashboard_domain_code}
\end{figure*}

\begin{figure*}[h!]
    \centering
    \includegraphics[width=0.9\linewidth, trim=1.4cm 6.2cm 1.4cm 0cm, clip]{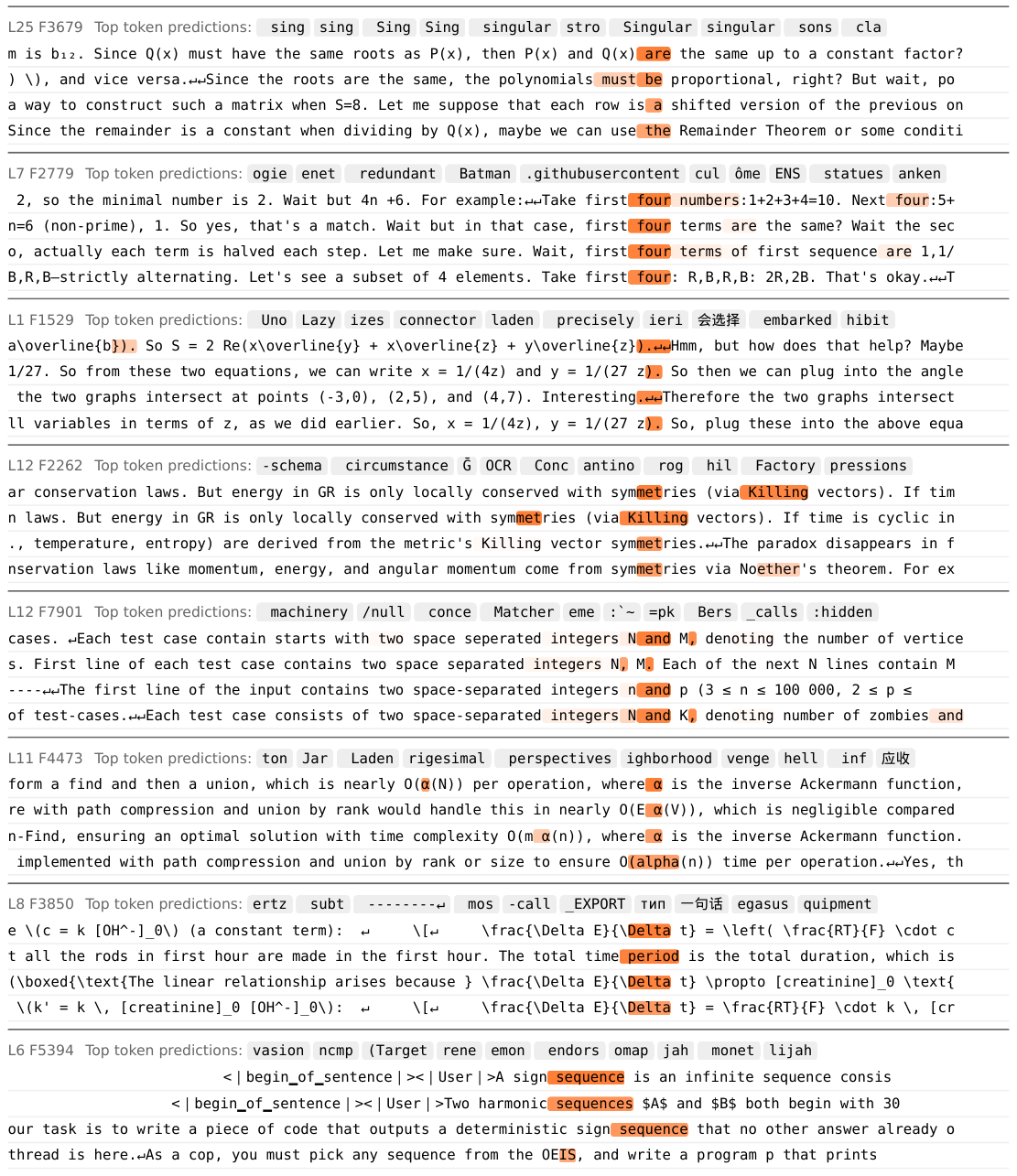}
    \caption{\textbf{Domain-specific features: math.} Randomly sampled transcoder adapter features classified by the LLM judge as domain-specific to math. For each feature, we show four max-activating dataset examples alongside the tokens most promoted by the feature's decoder direction.}
    \label{fig:dashboard_domain_math}
\end{figure*}

\begin{figure*}[h!]
    \centering
    \includegraphics[width=0.9\linewidth, trim=1.4cm 6.2cm 1.4cm 0cm, clip]{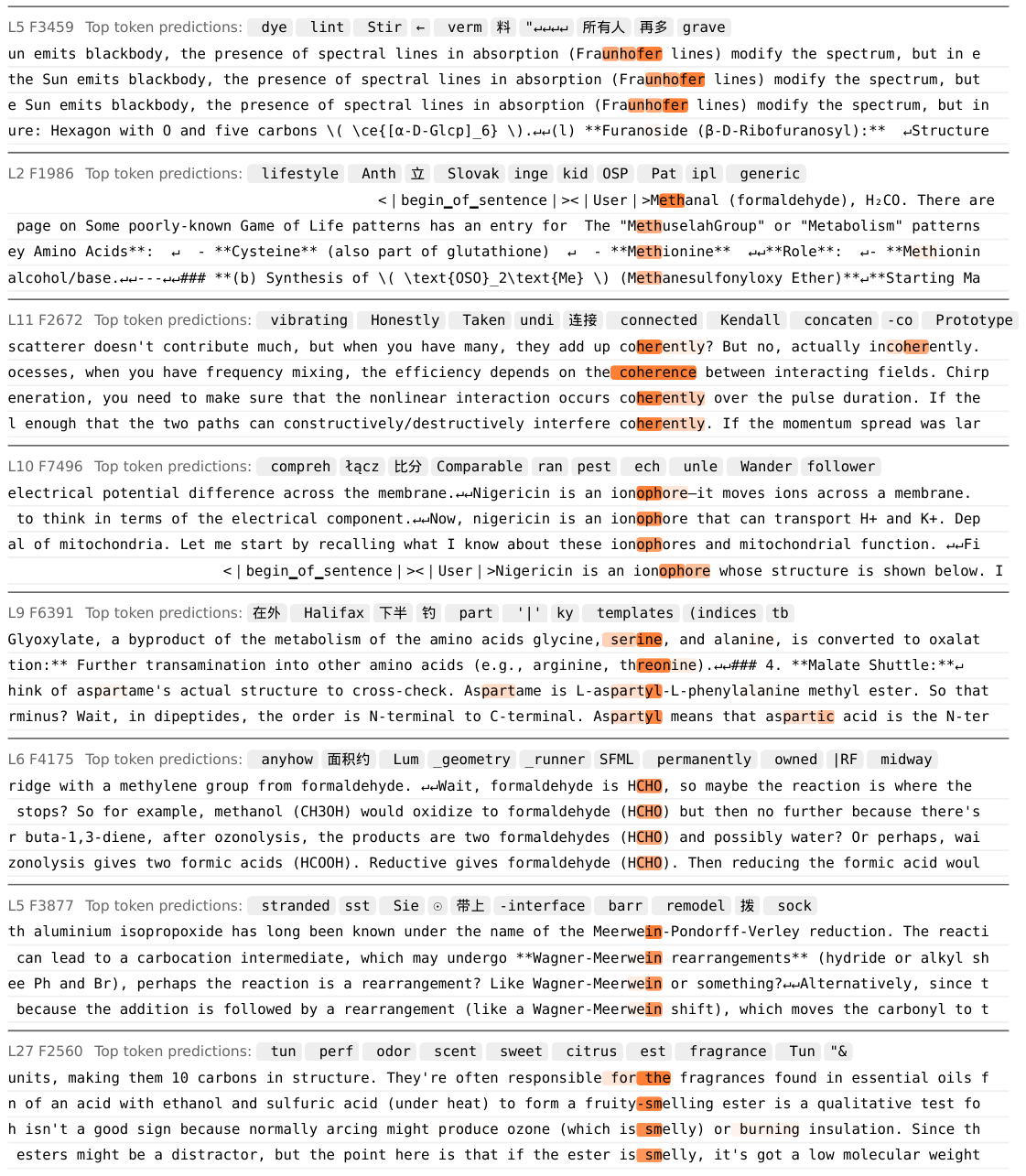}
    \caption{\textbf{Domain-specific features: science.} Randomly sampled transcoder adapter features classified by the LLM judge as domain-specific to science. For each feature, we show four max-activating dataset examples alongside the tokens most promoted by the feature's decoder direction.}
    \label{fig:dashboard_domain_science}
\end{figure*}

\begin{figure*}[h!]
    \centering
    \includegraphics[width=0.9\linewidth, trim=1.4cm 6.2cm 1.4cm 0cm, clip]{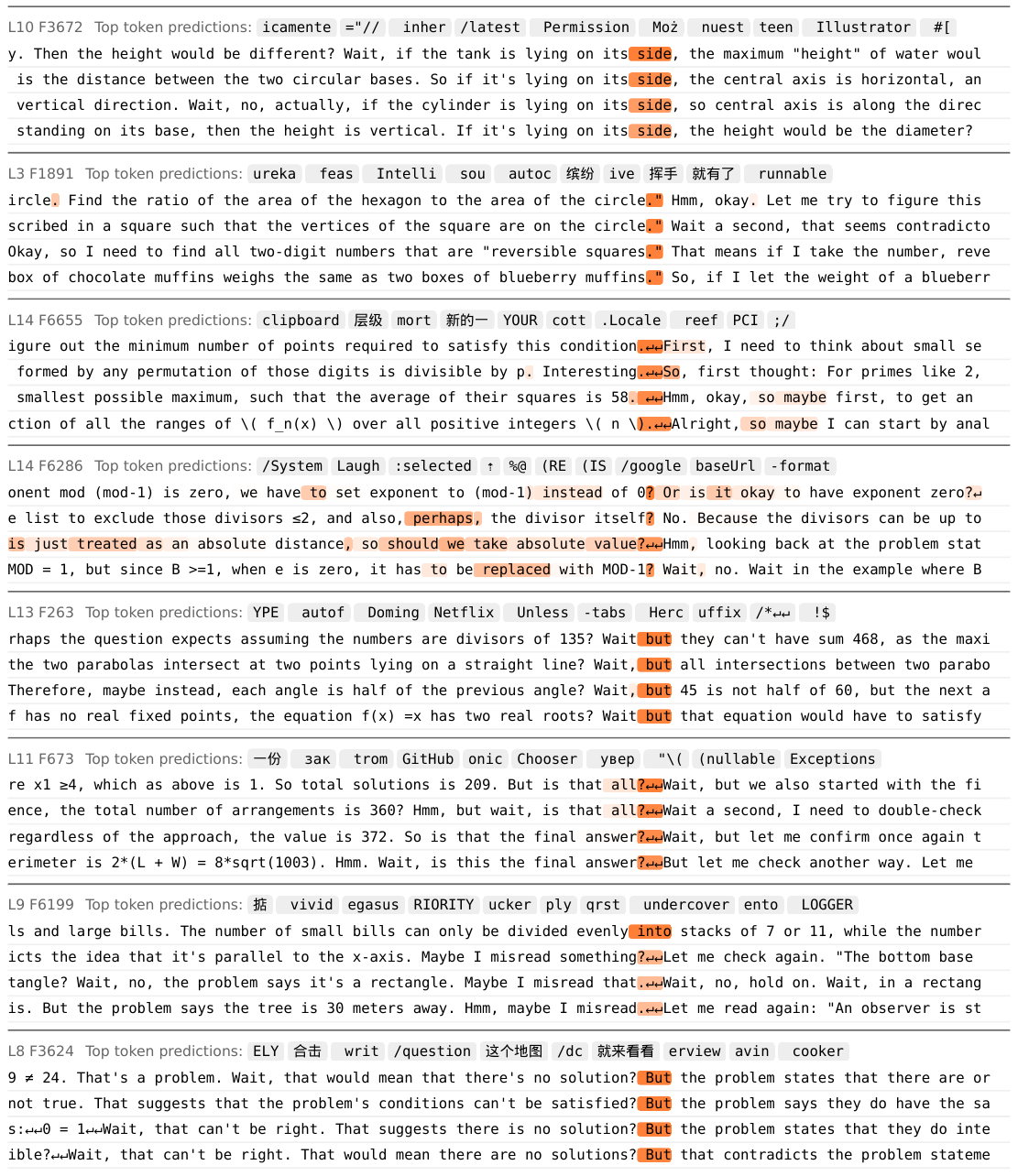}
    \caption{\textbf{Reasoning-related features: abstract input.} Randomly sampled transcoder adapter features classified by the LLM judge as reasoning-related with abstract input patterns. For each feature, we show four max-activating dataset examples alongside the tokens most promoted by the feature's decoder direction.}
    \label{fig:dashboard_input_abstract}
\end{figure*}

\begin{figure*}[h!]
    \centering
    \includegraphics[width=0.9\linewidth, trim=1.4cm 6.2cm 1.4cm 0cm, clip]{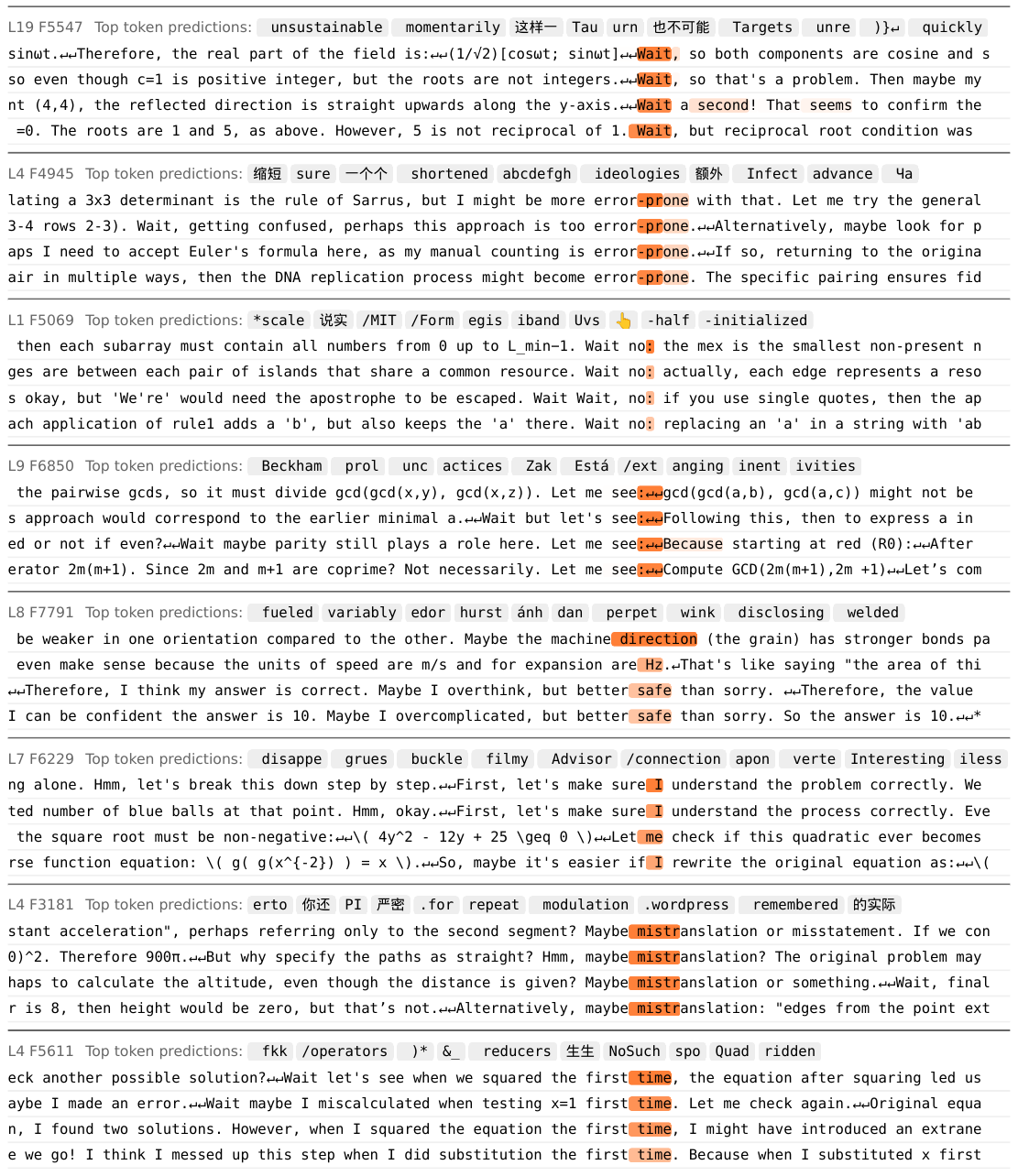}
    \caption{\textbf{Reasoning-related features: simple input.} Randomly sampled transcoder adapter features classified by the LLM judge as reasoning-related with simple input patterns. For each feature, we show four max-activating dataset examples alongside the tokens most promoted by the feature's decoder direction.}
    \label{fig:dashboard_input_simple}
\end{figure*}

\begin{figure*}[h!]
    \centering
    \includegraphics[width=0.9\linewidth, trim=1.4cm 6.2cm 1.4cm 0cm, clip]{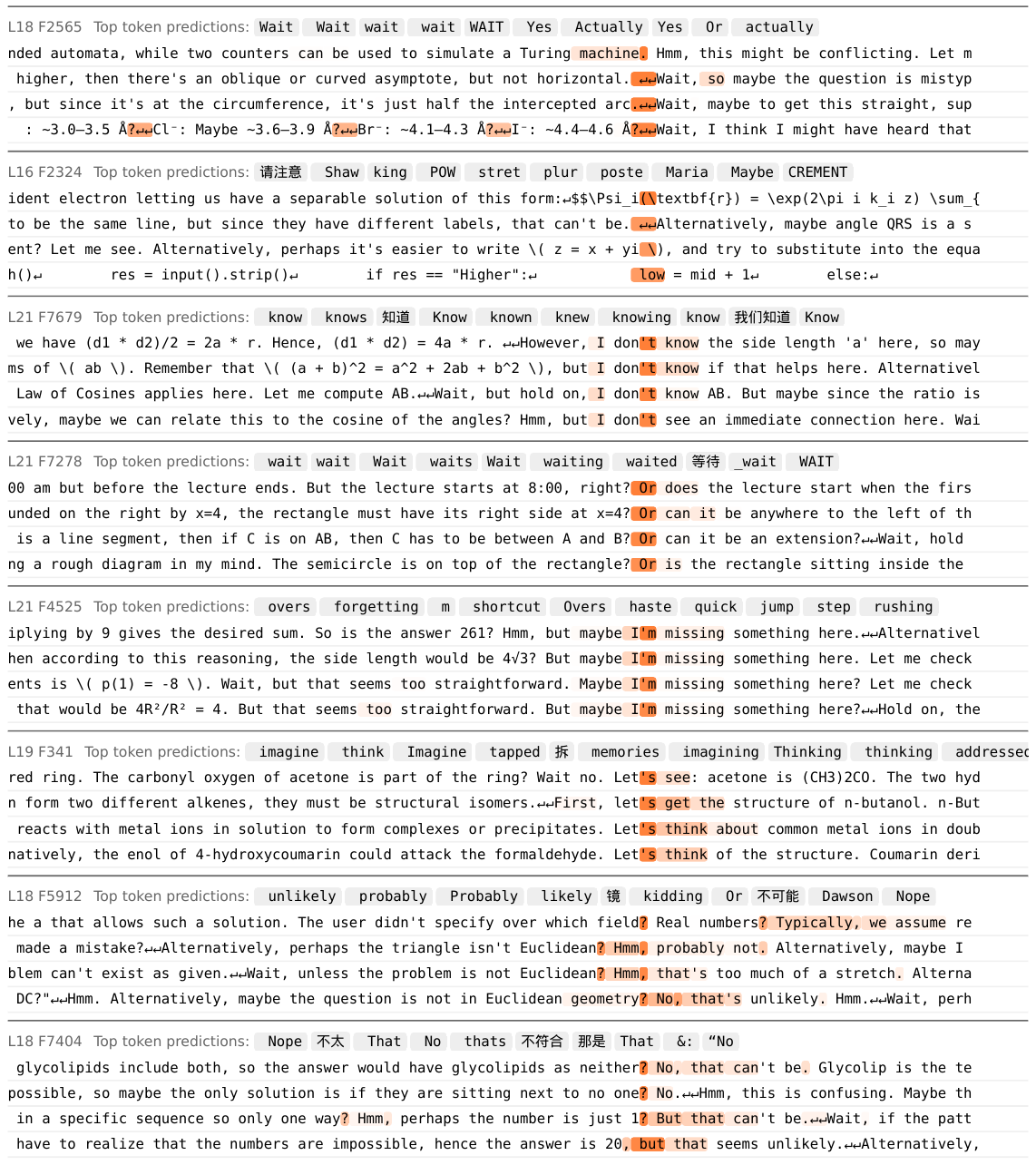}
    \caption{\textbf{Reasoning-related features: output.} Randomly sampled transcoder adapter features classified by the LLM judge as reasoning-related output features. For each feature, we show four max-activating dataset examples alongside the tokens most promoted by the feature's decoder direction.}
    \label{fig:dashboard_output}
\end{figure*}

\begin{figure*}[h!]
    \centering
    \includegraphics[width=0.9\linewidth, trim=1.4cm 6.2cm 1.4cm 0cm, clip]{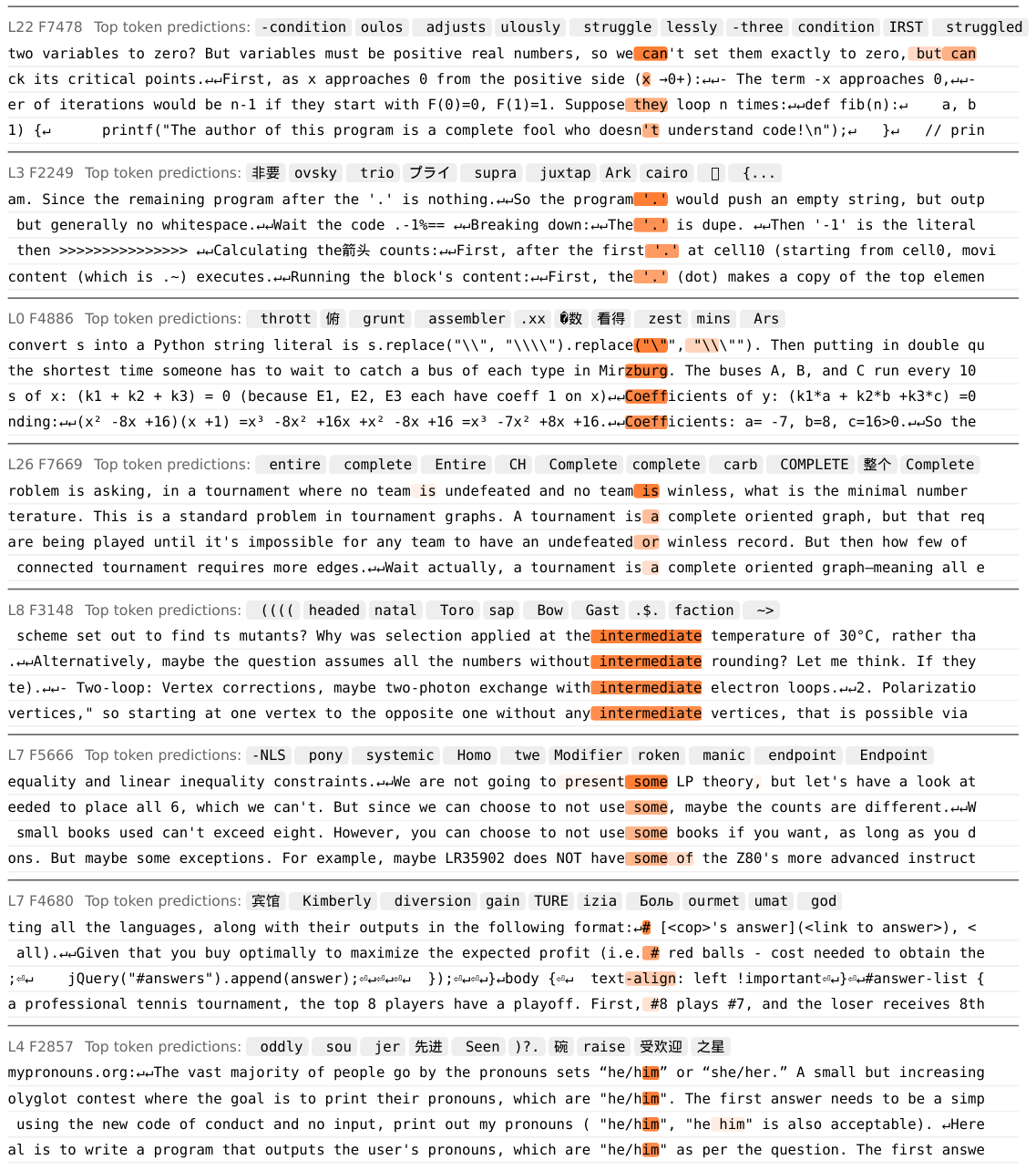}
    \caption{\textbf{Language modeling features.} Randomly sampled transcoder adapter features classified by the LLM judge as general language modeling features. For each feature, we show four max-activating dataset examples alongside the tokens most promoted by the feature's decoder direction.}
    \label{fig:dashboard_language}
\end{figure*}

\begin{figure*}[h!]
    \centering
    \includegraphics[width=0.9\linewidth, trim=1.4cm 6.2cm 1.4cm 0cm, clip]{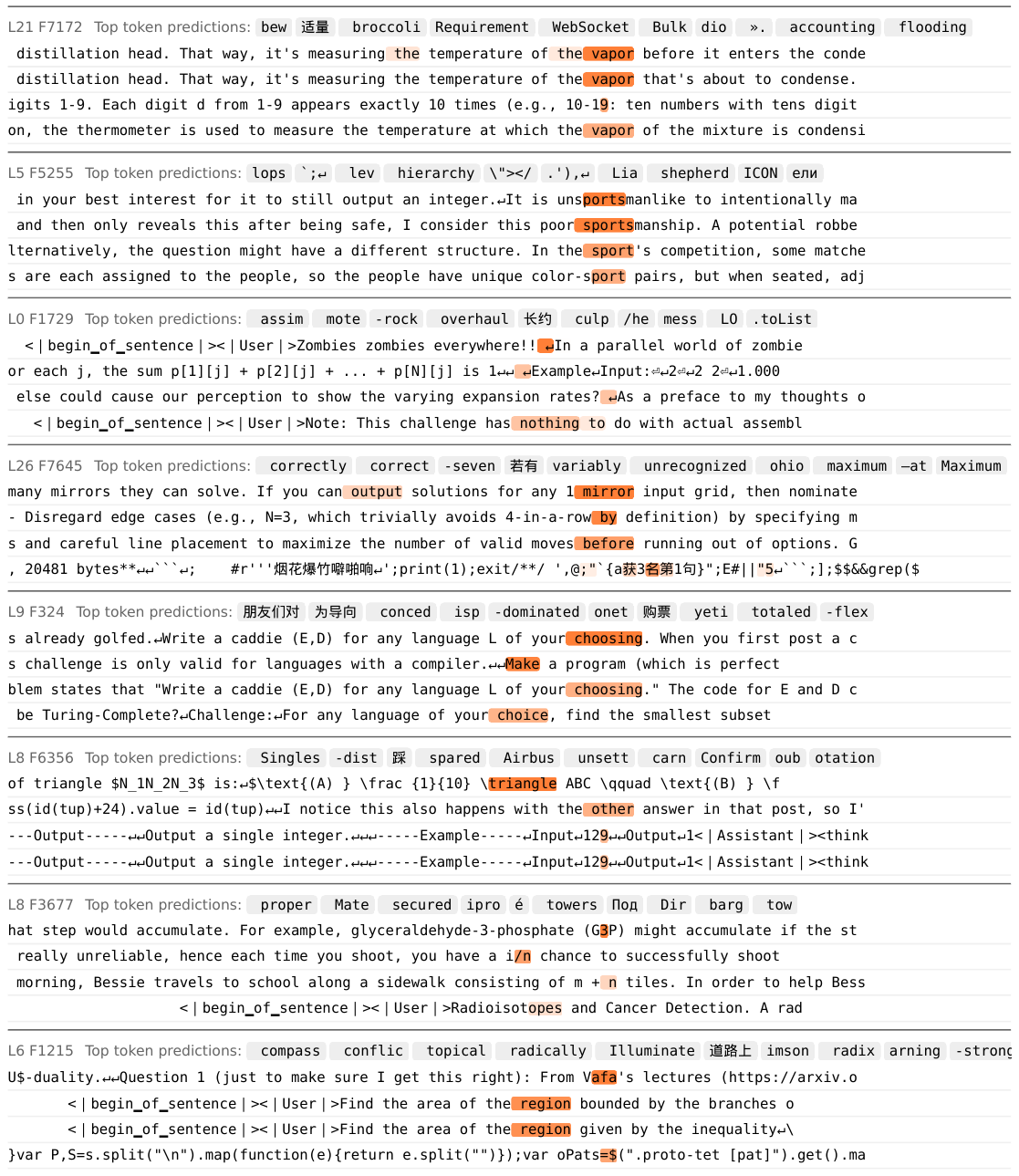}
    \caption{\textbf{Uninterpretable features.} Randomly sampled transcoder adapter features deemed uninterpretable by the LLM judge. For each feature, we show four max-activating dataset examples alongside the tokens most promoted by the feature's decoder direction.}
    \label{fig:dashboard_uninterpretable}
\end{figure*}

\end{document}